\documentclass[10pt,twocolumn,letterpaper]{article}

\usepackage[pagenumbers]{cvpr} 

\usepackage{amsmath}
\usepackage{graphicx}
\usepackage{xspace}

\usepackage{amsfonts}

\usepackage{tikz}
\usetikzlibrary{arrows.meta, positioning}
\usepackage{booktabs}
\usepackage{multirow, tabularx}
\usepackage[detect-all,per-mode=symbol]{siunitx}
\usepackage{makecell}
\usepackage{float}
\usepackage{subcaption}

\definecolor{cvprblue}{rgb}{0.21,0.49,0.74}
\usepackage[pagebackref,breaklinks,colorlinks,allcolors=cvprblue]{hyperref}

\title{On\,the\,Role\,of\,Rotation\,Equivariance\,in\,Monocular\,2D-to-3D\,Human\,Pose\,Lifting
\vspace{-28pt}}

\author{Pavlo Melnyk$^{1}$ \quad Cuong Le$^{1}$ \quad Urs Waldmann$^{1}$ \quad Per-Erik Forssén$^{1}$ \quad Bastian Wandt$^{2}$\\
{$^{1}$CVL, Linköping University  \quad $^{2}$Independent Researcher}\\
{\tt\small pavlo.melnyk@liu.se \quad bastianwandt@gmail.com}
\vspace{-12pt}
}

\begin{document}
\maketitle
\begin{abstract}
We consider monocular 3D human pose estimation (HPE), focusing on lifting detected 2D keypoints to 3D skeletal joints.
Despite their success, we find that commonly used lifting models, including a state-of-the-art probabilistic method, exhibit substantial performance degradation under in-plane rotations.
We systematically study SO(2) rotation equivariance and robustness across lifting models, comparing approaches that learn approximate equivariance via data augmentation with those that enforce exact equivariance via architectural constraints or pose canonicalisation.
We investigate their effects on prediction accuracy, rotational robustness, cross-dataset generalisation, and computational efficiency.
Using established HPE benchmarks, we show that rotation-based augmentation applied jointly to input and output poses substantially improves accuracy on rotated poses, reducing error by up to 72\% with little loss of original-pose accuracy.
These improvements also extend to probabilistic lifting and real-world sequences containing natural full-body rotations.
While augmented models can outperform specialised equivariant architectures at considerably lower inference cost, comparisons using canonicalisation-based versions of the same lifting models demonstrate that exact equivariance need not compromise prediction accuracy.
Direct equivariance diagnostics further show that rotational robustness does not imply exact equivariance.
Overall, the relative benefits of learning or enforcing equivariance depend on the lifting backbone and the mechanism used to achieve it.
\end{abstract}    
\vspace{-10pt}
\section{Introduction}
\label{sec:intro}
\vspace{-3pt}
\begin{figure}[t]
    \centering
    \includegraphics[width=0.8\linewidth]{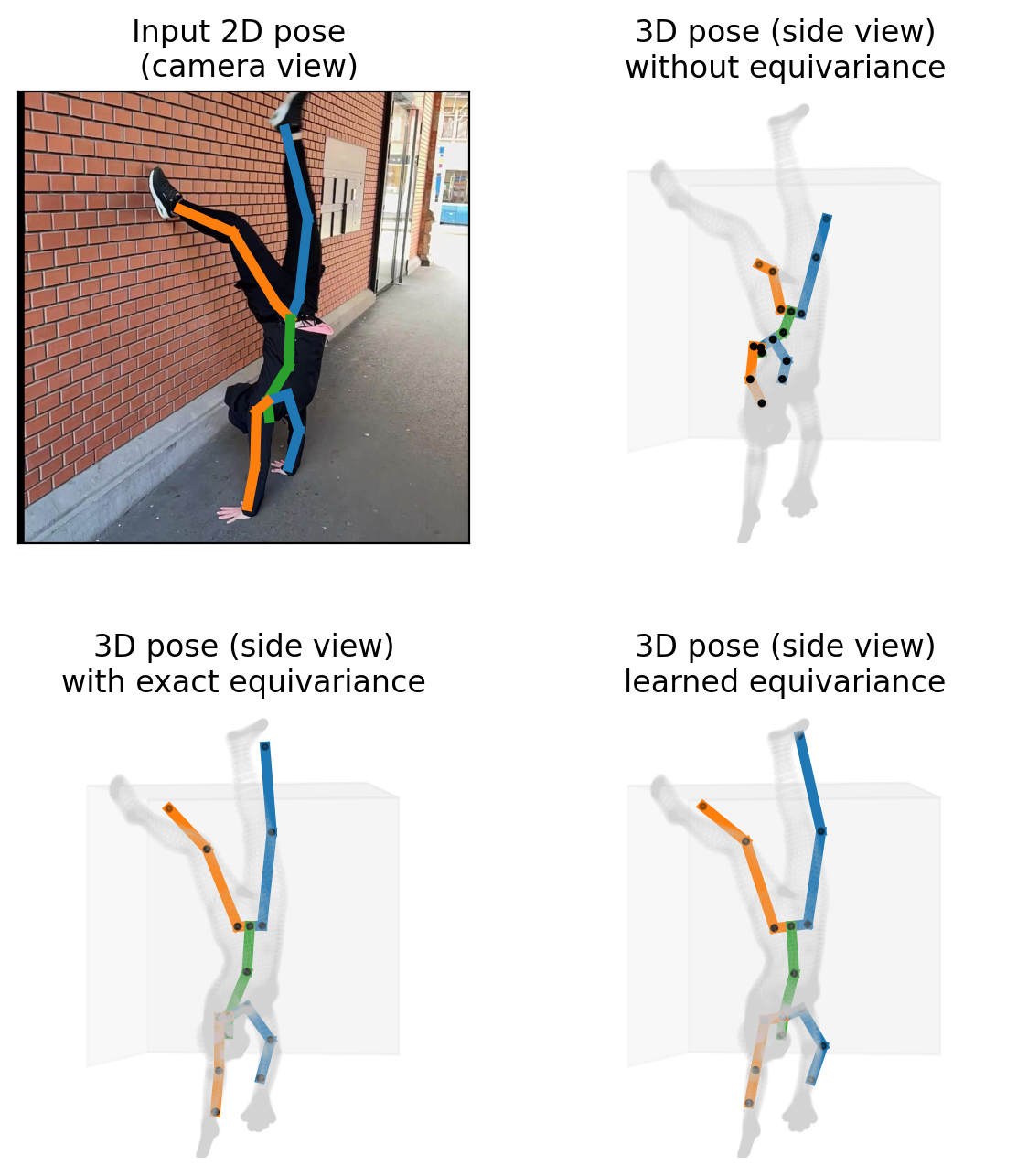}
    \caption{
    Typical 2D-to-3D lifting models for monocular human pose estimation exhibit substantial performance degradation under in-plane rotations.
    Learning approximate rotation equivariance via data augmentation substantially improves accuracy on such poses. 
    Exact SO(2) equivariance can be enforced by design either architecturally or using pose canonicalisation. 
    The resulting accuracy depends on the lifting model and the mechanism used to achieve equivariance. 
    Blue, green, and orange encode the right, middle, and left parts of the human body, respectively. The ground-truth 3D pose is shown in grey. 
    The sample is from the \texttt{68\_outdoor\_handstand} sequence of EMDB~\cite{kaufmann_emdb}.}
    \vspace{-15pt}
    \label{fig:teaser}
\end{figure}
In this work, we focus on monocular 3D human pose estimation (HPE), where the goal is to estimate 3D human skeletal joints from a single 2D image.
This problem is inherently ill-posed due to the ambiguity introduced by single-view (\ie monocular) observations.

While there exist methods that solve this task in an end-to-end manner, a common approach is two-stage: 1) detect 2D keypoints from the input image, \eg by using a pre-trained detector such as HRNet~\cite{cheng20_hrnet}, and 2) lift the detected 2D keypoints to 3D by means of another trained model.
An important property to consider for 2D-to-3D lifting is SO(2) equivariance: rotating a normalised, pelvis-centred 2D pose should induce the corresponding rotation of the predicted 3D pose in the $xy$-plane, while leaving the depth coordinate unchanged.
Typical lifting models, however, do not guarantee this property, and may exhibit substantial prediction errors under rotations, \eg see Figure~\ref{fig:teaser}.

To address this, we investigate the practical benefits and limitations of enforcing exact SO(2) equivariance via architectural desing or pose canonicalisation versus learning approximate equivariance via rotation-based data augmentation.
Specifically, we examine whether exact equivariance necessarily compromises prediction accuracy, or whether the observed trade-offs depend on the mechanism used to enforce it.
We evaluate these approaches on the common human pose estimation benchmarks Human3.6M~\cite{ionescu_h36m} and MPI-INF-3DHP~\cite{dushyant_3dhp}, as well as EMDB~\cite{kaufmann_emdb} and SportsCap~\cite{chen_sportscap}. 
This combines controlled rotations of detected 2D poses and cross-dataset testing, as well as real-world sequences containing natural full-body rotations. 

The primary scope of our work is single-frame monocular 3D HPE, focusing on equivariance within the 2D-to-3D lifting stage using detected or ground-truth 2D keypoints, without additional image features or temporal information.

Our contributions are that we systematically investigate SO(2) equivariance and robustness to rotations in monocular 2D-to-3D human pose lifting, and quantitatively and qualitatively demonstrate that
\begin{itemize}
    \item Typical lifting models exhibit substantial accuracy degradation under in-plane rotations, while rotation-based data augmentation improves accuracy on rotated poses and selected naturally rotated sequences, substantially reducing equivariance error.
    \item Exact SO(2) equivariance does not necessarily compromise prediction accuracy.
    By introducing canonicalisation-based versions of representative lifting models, we enable controlled comparisons between learning approximate equivariance and enforcing exact equivariance using the same lifting backbones.
    \item The effectiveness of equivariance depends on the lifting backbone and the mechanism used to achieve it.
    Across different types of lifting models, we show that rotational robustness alone does not establish exact equivariance, and that canonicalisation can achieve competitive accuracy at relatively low computational cost.
\end{itemize}

\vspace{-5pt}
\section{Related Work and Background}
\label{sec:background}
\vspace{-4pt}
\subsection{Human Pose Estimation}
\vspace{-4pt}
Monocular 3D human pose estimation is a broad and challenging field of research.
Methods can be split into two categories: two-stage and end-to-end approaches.
\\\noindent\textbf{Two-stage approaches}\quad
Two-stage approaches decompose monocular 3D human pose estimation into two stages: 2D keypoint detection and 3D pose lifting. This allows the lifting module to focus completely on geometric reasoning.

A well-established baseline in this category is the work by Martinez~\etal~\cite{martinez17_3dbaseline}, which uses a lightweight fully connected network to regress 3D joint coordinates from 2D inputs. Despite its simplicity, this method achieves strong performance on Human3.6M~\cite{ionescu_h36m} and clearly demonstrates that accurate 3D pose estimation is feasible given reliable 2D keypoints.
Due to its effectiveness and minimal architectural assumptions, it remains a standard baseline for evaluating lifting-based methods, including in our work.

Subsequent research extends this paradigm by incorporating additional constraints and refinement strategies.
Complementary to fully supervised approaches, Wandt and Rosenhahn~\cite{Wandt_2019_CVPR} propose RepNet, a weakly supervised lifting framework that combines adversarial training with camera estimation and reprojection constraints to learn 3D poses from 2D keypoints.
Li~\etal~\cite{Li_2020_CVPR} propose a cascaded lifting framework with iterative refinement and data evolution to improve robustness to noisy 2D detections.
Xu~\etal~\cite{Xu_2020_CVPR} integrate kinematic priors directly into the lifting network, enforcing physically plausible joint relationships to reduce depth ambiguity and implausible poses.
PoseAug~\cite{gong2021poseaug} introduces a differentiable pose augmentor that jointly learns to modify posture, body size, viewpoint, and position to improve the generalisation of 2D-to-3D lifting models.
Transformer-based architectures have also been adopted to enhance lifting performance by modelling global spatial and temporal dependencies: PoseFormer~\cite{zheng21_poseformer} 
introduces a spatiotemporal transformer architecture to model dependencies between joints and across frames for 3D human pose estimation.
Extending lifting methods to more challenging viewpoints, Wang~\etal~\cite{Wang_2023_CVPR} augment 2D keypoints with scene geometry cues, addressing severe occlusions and missing joints common in egocentric settings.

More recently, probabilistic lifting methods have been proposed to account for the inherent ambiguity of monocular 3D HPE.
Holmquist and Wandt~\cite{Holmquist_2023_ICCV} introduce DiffPose, a conditional diffusion model that generates multiple plausible 3D pose hypotheses from sampled 2D keypoint candidates.
Le~\etal~\cite{le2026flow} propose FMPose, using flow matching to model the conditional distribution of 3D poses given 2D observations.
Similarly, Wang~\etal~\cite{Wang_2026_CVPR} introduce FMPose3D, combining flow matching with reprojection-based aggregation of multiple 3D hypotheses.
\\\noindent\textbf{End-to-end approaches}\quad
End-to-end methods aim to predict 3D human pose directly from RGB images, eliminating the explicit 2D lifting stage.
Wang~\etal~\cite{Wang_2022_CVPR} introduce a unified framework that jointly performs detection and 3D pose estimation while modelling uncertainty through distribution-aware representations, enabling robust multi-person inference in crowded scenes.
Earlier work from Moon~\etal~\cite{Moon_2019_ICCV} incorporates camera distance estimation into a top-down pipeline to alleviate scale and depth ambiguities inherent in monocular images.
PersPose~\cite{hao2025perspose} incorporates camera-intrinsics encoding and perspective rotation of input images to reduce perspective distortion and improve monocular 3D pose estimation.
\subsection{Rotation Equivariance}
\vspace{-4pt}
Rotations in $n$D are the actions of the special orthogonal group $\text{SO}(n)$, and can be represented by $n \times n$ matrices $R$ such that $R^\top R = R R^\top = \textup{I}_n$, with $\textup{I}_n$ being the identity matrix, and $\det{R} = 1$.

A function $f : \mathcal{X} \rightarrow \mathcal{Y}$ is said to be $\text{SO}(n)$-equivariant, \ie equivariant under $n$D rotations, if for every $R \in \text{SO}(n)$, in the function output space, $\mathcal{Y}$, there exists a transformation representation, $\rho(R)$, such that
\vspace{-3pt}
\begin{equation}
\label{eq:equivariance}
    \rho(R) \, f(X) = f(R X) \text{\quad for all~} R \in \text{SO}(n), \; X \in \mathcal{X} \subseteq \mathbb{R}^n.
\end{equation}
If the transformation representation in the output space is the identity, \ie if for every  $R \in \text{SO}(n)$, $\rho(R) = \textup{I}_n$, we call a function $f: \mathcal{X} \rightarrow \mathcal{Y}$~~$\text{SO}(n)$-invariant:
\vspace{-3pt}
\begin{equation}
\label{eq:invariance}
    f(X) = f(R X) \text{\quad for all~} R \in \text{SO}(n), \; X \in \mathcal{X} \subseteq \mathbb{R}^n.
\end{equation}
Rotation equivariance is a powerful inductive bias, and when incorporated appropriately, leads to improved and robust performance, \eg in point-cloud classification \cite{deng2021vector}, trajectory prediction \cite{assadd23_vntransformer}, molecular properties prediction \cite{aykent25_gotennet}, and energy and force prediction for materials science applications \cite{Equiformerv2}.

Gerken~\etal~\cite{gerken2022equivariance} compare architectural equivariance with rotation augmentation for spherical images, finding that their relative benefits depend on the task, with equivariant networks outperforming augmented alternatives in semantic segmentation.
V\'eges~\etal~\cite{veges20193d} propose a siamese architecture that learns rotation-equivariant embeddings for 2D-to-3D human pose lifting, aiming to improve generalisation to unseen camera viewpoints while reducing reliance on data augmentation.
Kaba~\etal~\cite{kaba2023equivariance} propose learned canonicalisation functions that can be combined with non-equivariant backbones to construct equivariant models.
In contrast, we use parameter-free pose canonicalisation to enable controlled comparisons of equivariance mechanisms using the same lifting backbones.
Howell~\etal~\cite{howell2023equivariant} study equivariant 3D pose prediction from 2D images, showing that geometric consistency imposes SO(2)-equivariance constraints associated with rotations about the viewing direction.
DECA~\cite{garau2021deca} uses capsule autoencoders to improve viewpoint equivariance and generalisation in human pose estimation from RGB and depth images.

In our work, we focus specifically on SO(2) equivariance in 2D-to-3D human pose lifting, using 2D keypoints as input.
We systematically compare rotation augmentation, specialised equivariant architectures, and pose canonicalisation, examining their effects on prediction accuracy, rotational robustness, and computational efficiency.
\vspace{-4pt}
\section{Methodology}
\vspace{-4pt}
\subsection{SO(2) Equivariance in Lifting Models}
\label{sec:geometrically_consistent}
\vspace{-4pt}
\label{sec:method}
We consider 2D-to-3D lifting from detected 2D joint coordinates rather than directly from images. 
Let $P\in\mathbb{R}^{N\times2}$ denote the detected 2D pose and $\mathbf{X}\in\mathbb{R}^{N\times3}$ the corresponding root-relative 3D pose in camera coordinates. 
Before lifting, we normalise each 2D pose by subtracting the mean joint position, dividing by a single scalar standard deviation shared across both axes, and centring the result at the pelvis. 
We denote the resulting 2D pose by $X\in\mathbb{R}^{N\times2}$.
This isotropic normalisation preserves the action of in-plane rotations.
We study equivariance under global rotations ($R_\theta\in \text{SO}(2)$) applied to these normalised, pelvis-centred 2D coordinates. 
The derivation of this rotation-preserving normalisation is provided in Supplementary~Section~\ref{supp:normalisation}.

\begin{figure}[t]
    \centering
    \includegraphics[width=\linewidth]{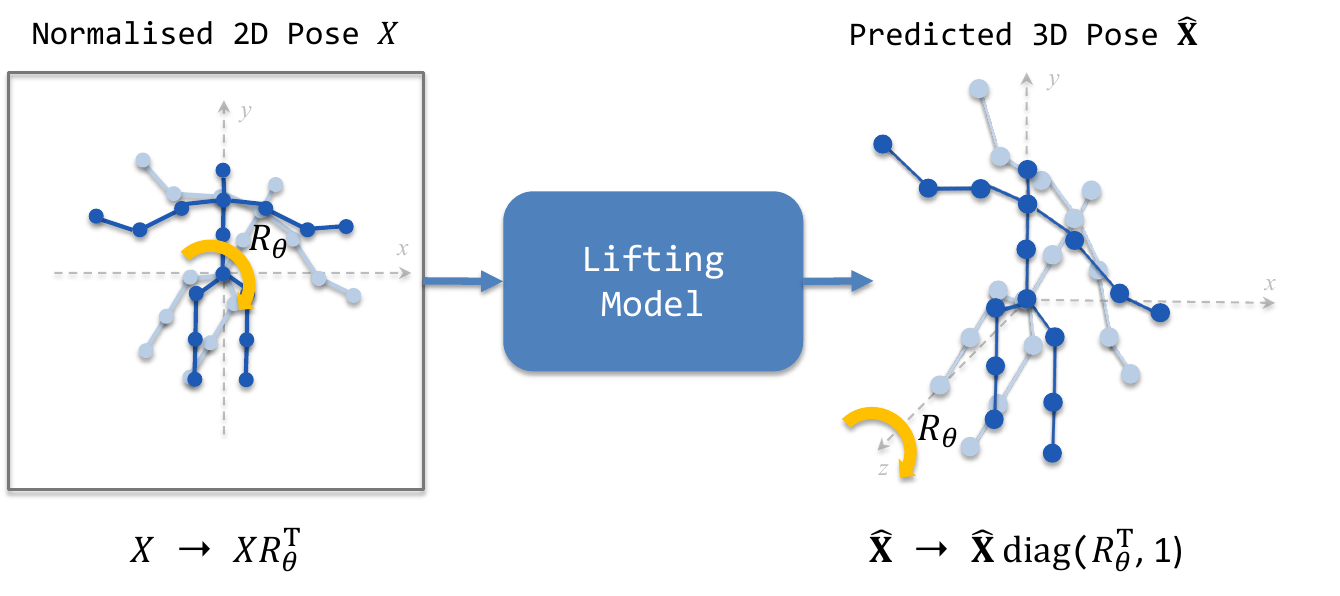}
    \caption{
     SO(2)-equivariance in 2D-to-3D human pose lifting. 
     A global rotation of the normalised, pelvis-centred 2D pose induces the corresponding rotation of the predicted root-relative 3D pose in the $xy$-plane while leaving depth unchanged.}
    \label{fig:geometric_consistency}
    \vspace{-12pt}
\end{figure}
We call a lifting model, $f: \mathbb{R}^{N\times2} \rightarrow \mathbb{R}^{N\times3}$, SO(2)-equivariant if the following equivariance property holds for any $X \in \mathbb{R}^{N\times2}$ and an in-plane rotation, $R_\theta \in \text{SO}(2)$, by an angle $\theta \in [0, 2\pi)$:
\begin{equation}
\label{eq:geometrical_consistency}
    f(X\,R^\top_\theta) =
    f(X) \begin{bmatrix}
        R^\top_\theta & 0 \\
        0 & 1
    \end{bmatrix} ,
\end{equation}
which means that a global in-plane rotation of the normalised, pelvis-centred 2D pose induces the corresponding rotation of the predicted 3D $xy$-coordinates, while leaving depth invariant (see Figure~\ref{fig:geometric_consistency}).
The interpretation of the SO(2) action as a physical rotation about the camera’s optical axis additionally requires appropriately calibrated, isotropically scaled image coordinates.

In Section~\ref{sec:experiments}, we systematically study how the SO(2)-equivariance constraint in Eq.~\eqref{eq:geometrical_consistency} affects 2D-to-3D pose lifting, distinguishing between rotational robustness learned via data augmentation and equivariance enforced by design. 
We further investigate whether the mechanism used to enforce equivariance influences model accuracy, generalisation, and computational efficiency by employing the following types of models:
\begin{itemize}
    \item \textbf{(Fully) Equivariant}: Models that satisfy Eq.~\eqref{eq:geometrical_consistency} exactly by design, either via equivariant internal representations (\textbf{architecture-based}) or via deterministic input canonicalisation and inverse output transformation (\textbf{canonicalisation-based})
    \item \textbf{Hybrid}: Models that only produce rotation-equivariant $xy$-coordinates by architecture design, with depth without an invariance guarantee (\textbf{hybrid}) and trained to produce approximately invariant depth (\textbf{hybrid\,+\,aug})
    \item \textbf{Vanilla}: Models that are non-equivariant (\textbf{vanilla}) and trained to approximately satisfy Eq.~\eqref{eq:geometrical_consistency} by augmentation (\textbf{vanilla\,+\,aug}).
\end{itemize}
\subsection{Models}
\label{sec:models}
\vspace{-3pt}
In this section, we outline the models we build according to the types described above.
We strive to conduct a fair comparison and therefore construct the models with a similar number of parameters (${\sim}3.7-4.5$M) by matching, when applicable, the number of layers and feature dimensions.
Our choice of models is motivated by their relevance to the scope of our work (see Section~\ref{sec:intro}), reported performance on related tasks, and reproducibility.
\begin{figure}[t]
\centering
\resizebox{\linewidth}{!}{%
\begin{tikzpicture}[
  x=\linewidth,
  y=2.4cm,
  font=\small,
  box/.style={draw, rounded corners, minimum height=10mm, minimum width=48mm, align=center},
  arrow/.style={->, thick},
  dottedline/.style={dotted, thick}
]

\node (in) at (0.0,0) {$X \in \mathbb{R}^{N\times 2}$};

\coordinate (inTop) at (0.10, 0.35);
\coordinate (inBot) at (0.10,-0.35);

\node[box] (eq)   at (0.48, 0.35) {2D Equivariant Model};
\node[box] (base) at (0.48,-0.35) {Non-equivariant Baseline};

\node (outxy) at (0.96, 0.35) {$xy$-coordinates};
\node (outz)  at (0.96,-0.35) {$z$-coordinate};
\node (out) at (0.96,0) {$\hat{\mathbf{X}} \in \mathbb{R}^{N\times 3}$};

\draw[arrow] (inTop) -- (eq.west);
\draw[arrow] (inBot) -- (base.west);

\draw[arrow] (eq.east) -- (outxy.west);
\draw[arrow] (base.east) -- (outz.west);

\draw[dottedline]
  (eq.south)
  -- (0.48,0)      
  -- (0.10,0)      
  -- (inBot);      

\end{tikzpicture}
}
\caption{\textbf{Hybrid} model architecture outline: the $xy$-outputs are rotation-equivariant, whereas the $z$-coordinate prediction is non-equivariant.
In the default setting (no dotted line), the input is fed into both models in parallel; for the ablation study (with the dotted line), the input to the non-equivariant baseline is the output of the first equivariant layer of the 2D equivariant model.}
\vspace{-12pt}
\label{fig:hybrid_model_types}
\end{figure}
\vspace{-4pt}
\subsubsection{Equivariant}
\vspace{-3pt}
\noindent \textbf{Architecture-based} We utilise VN-Transformer~\cite{assadd23_vntransformer}, a transformer architecture based on the vector neurons framework~\cite{deng2021vector} for rotation-equivariant point cloud processing. 
In addition, we employ GotenNet~\cite{aykent25_gotennet}, a recent efficient rotation-equivariant model for molecular analysis.
We use their open-source implementations.
In order to utilise these models, we need to represent the input ($N \times 2$) and output ($N \times 3$) joints with the same dimensionality. 
While we can use a constant value as the third coordinate for all input points for VN-Transformer (\eg $1$), this does not work for GotenNet due to the joint-wise differences used in the model, which leads to the third coordinate in the output always being zero. 
Thus, for both models, we append to each 2D $xy$-vector in the $N\times 2$ input its L2 norm (which is rotation-invariant), \ie $z = \sqrt{x^2 + y^2}$, preserving rotation equivariance.
We present an ablation study on this in Supplementary~Section~\ref{supp:ablations}.
Although both models are permutation-equivariant with respect to their input tokens, both architectures explicitly encode joint identities: 
VN-Transformer uses learned joint-index embeddings, and GotenNet uses separate learned embeddings for each joint and its neighbours. 
Since these identity features are invariant under SO(2) actions, they preserve the rotation equivariance of the architectures.

\noindent \textbf{Canonicalisation-based} 
We construct fully SO(2)-equivariant predictors from the vanilla backbones (described in the section below), following the general canonicalisation approach of Kaba~\textit{et~al.}~\cite{kaba2023equivariance}.
Given a 2D pose $X\in\mathbb{R}^{N\times2}$, we compute a parameter-free rotation $R_C(X)\in \text{SO}(2)$ that aligns an anatomical reference vector $v$ with the $y$-axis, \ie $v \, R_C = (0, \lVert v \rVert)$. 
Specifically, we use the pelvis-to-neck vector $v_1=X_{\text{neck}}-X_{\text{pelvis}}$; if $v_1=0$, we instead use $v_2$, the vector from the hip midpoint to the shoulder midpoint.
We then apply the vanilla backbone $f$ to the canonicalised 2D pose and apply the inverse rotation to the output $xy$-coordinates with the depth unchanged:
\begin{equation}
    \label{eq:canonicalisation}
    \hat{\mathbf{X}} = f(X\,R_C(X))\begin{bmatrix}
        R_C(X)^\top & 0\\
        0 & 1
\end{bmatrix}.
\end{equation}
Both reference vectors transform consistently under in-plane rotations, and thus, the resulting model satisfies Eq.~\eqref{eq:geometrical_consistency} whenever at least one reference vector is non-zero.
This canonicalisation has no learnable parameters.
We provide the proof, further implementation details, and reference-coverage diagnostics in Supplementary~Section~\ref{supp:canonicalisation}.
\vspace{-15pt}
\subsubsection{Vanilla}
\vspace{-3pt}
\label{sec:vanilla}
We use the simple ResNet-based baseline by Martinez~\emph{et al.}~\cite{martinez17_3dbaseline}, as well as a transformer-based baseline, PoseFormer, by Zheng~\emph{et al.}~\cite{zheng21_poseformer}.
These vanilla models do not enforce rotation equivariance in their 3D human pose estimates, and can exhibit degraded performance when encountering a rotated out-of-distribution pose, as seen in the example in Figure~\ref{fig:teaser}.
The architecture from Martinez~\emph{et al.}~\cite{martinez17_3dbaseline} consists of two sequential ResNet blocks, taking the flattened 2D pose as input vector, and outputs the corresponding 3D pose vector.
PoseFormer~\cite{zheng21_poseformer} models spatial and temporal dependencies between human joints via the transformer architecture.
Originally, PoseFormer is designed to take a sequence of 2D poses as input.
For a fair comparison, we modify the architecture to work with single-frame inputs by reducing the length of the input sequence to $1$ and decreasing the depth of the temporal transformer module to $2$, resulting in a comparable number of learnable parameters.
\\\noindent\textbf{Reference method (FMPose)}\quad For reference and a broader perspective, we also include FMPose~\cite{le2026flow}, a state-of-the-art flow-matching-based method for monocular 3D HPE, with $4.5$M learnable parameters in total. 
However, unlike the other models within the scope of our work, FMPose operates on multiple 2D poses (sets of keypoints), obtained from HRNet-predicted~\cite{cheng20_hrnet} keypoint heatmaps, and produces multiple 3D hypotheses. 
In Section~\ref{sec:results}, we therefore report both its deterministic (single hypothesis) and probabilistic (best of multiple hypotheses) evaluations.
\vspace{-3pt}
\subsubsection{Hybrid}
\vspace{-3pt}
Additionally, we combine an equivariant model, VN-Transformer or GotenNet, used for $xy$-prediction only, with the ResNet baseline~\cite{martinez17_3dbaseline} predicting the depth coordinate, as shown in Figure~\ref{fig:hybrid_model_types}.
Given the VN-Transformer implementation, we directly instantiate it for 2D, whereas the GotenNet implementation is optimised for 3D, and we run it as-is by initialising the third coordinate to $0$ and extracting the first two coordinates for each joint in the output pose.
In both cases, we reduce the number of parameters of the equivariant models to keep the same non-equivariant baseline ResNet, so that the total number of parameters falls within the range of the other models in our comparison.

\vspace{-3pt}
\section{Experiments and Results}
\label{sec:experiments}
\vspace{-3pt}
\subsection{Datasets}
\begin{figure}[t]
    \centering
    \includegraphics[width=0.95\linewidth]{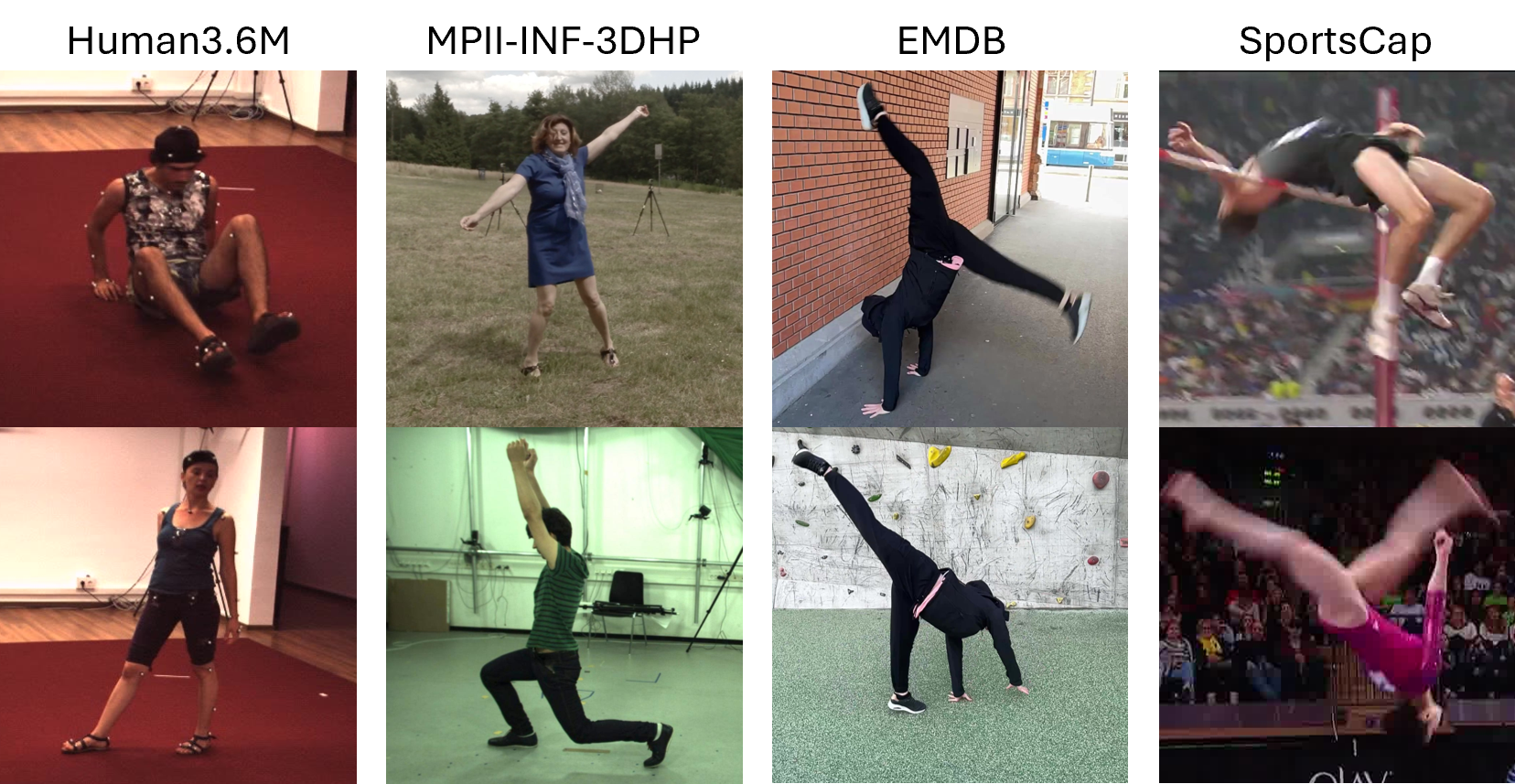}
    \caption{Samples from the datasets used in our experiments. The EMDB and SportsCap examples illustrate natural body rotations that are less represented in Human3.6M and MPI-INF-3DHP.}
    \vspace{-15pt}
    \label{fig:dataset}
\end{figure}

\vspace{-4pt}
We first train the models described in Section~\ref{sec:models} on the Human3.6M dataset~\cite{ionescu_h36m}, which contains diverse motion capture data in a laboratory setup.
Following the standard training protocol~\cite{martinez17_3dbaseline,zheng21_poseformer}, we use the first five subjects (S1, S5, S6, S7, S8) for training and the remaining two (S9, S11) for testing.
To examine the cross-dataset generalisation of the models, we additionally evaluate the models on the test set of the MPI-INF-3DHP dataset~\cite{dushyant_3dhp} without any fine-tuning.
We also evaluate the models on two EMDB~\cite{kaufmann_emdb} sequences, \texttt{68\_outdoor\_handstand} and \texttt{69\_outdoor\_cartwheel}, containing natural full-body rotations; see Figures~\ref{fig:teaser}~and~\ref{fig:qual_emdb}). 
We further provide qualitative results on representative samples from the SportsCap dataset~\cite{chen_sportscap} including complex rotated poses, for which 3D ground truth is unavailable (see Figure~\ref{fig:dataset}). 
\begin{table*}[t]
    \centering
    \setlength\tabcolsep{3pt}
    \footnotesize
        \caption{
    Quantitative results with evaluation \textit{Protocol~1} (MPJPE, $\downarrow$) on the \textit{Original} and \textit{Rotated} test sets of Human3.6M and MPI-INF-3DHP, as well as a subset of EMDB (the test sequences 
\texttt{68\_outdoor\_handstand} and \texttt{69\_outdoor\_cartwheel}) containing realistic full-body rotations.
    The best and second-best results among methods using single 2D pose as input are highlighted in \textbf{bold} and \underline{underlined}, respectively.
    *FMPose is included as a reference state-of-the-art method and differs from the other models by operating on multiple 2D pose samples (from a single image) and producing multiple 3D hypotheses; we report single (deterministic) and oracle best-of-hypotheses (probabilistic); see Section~\ref{sec:vanilla} for details; the \textbf{best}  and \underline{second-best} results for FMPose are highlighted separately from the other models.
    $^\dagger$ Evaluations using ground-truth 2D keypoints instead of HRNet detections.
    }
    \begin{tabularx}{\linewidth}{@{\extracolsep{\fill}}llccccccc}
        \toprule
        \multirow{2}{*}{Method} & \multirow{2}{*}{Type} & \multicolumn{2}{c}{Human3.6M} & \multicolumn{2}{c}{MPI-INF-3DHP} & \multicolumn{2}{c}{MPI-INF-3DHP$^\dagger$} & EMDB$^\dagger$ \\
        & & \textit{Original} & \textit{Rotated} & \textit{Original} & \textit{Rotated} & \textit{Original} & \textit{Rotated} &  \textit{Original}\\
        \midrule
        ResNet~\cite{martinez17_3dbaseline} & Vanilla & $\textbf{62.9}\pm{\textbf{0.2}}$ & $209.6\pm{1.4}$ & $144.0\pm{0.7}$ & $225.8\pm{2.9}$ & $109.1\pm3.7$ & $210.6\pm3.5$ & $155.3\pm1.2$ \\
        ResNet+aug & Vanilla & ${64.2}\pm{{0.1}}$ & $\underline{64.3}\pm{\underline{0.1}}$ & $152.3\pm{1.1}$ & $152.6\pm{1.4}$ & $\underline{108.5}\pm\underline{1.6}$ & $\underline{107.2}\pm\underline{1.1}$ & $\underline{114.3}\pm\underline{1.7}$ \\
        PoseFormer~\cite{zheng21_poseformer} & Vanilla & $65.9\pm{0.1}$ & $232.4\pm{5.2}$ & $\textbf{139.9}\pm{\textbf{0.7}}$ & $247.5\pm5.7$ & $118.7\pm2.3$ & $237.0\pm5.1$ & $167.5\pm2.9$ \\
        PoseFormer+aug & Vanilla & $65.1\pm{0.1}$ & $65.0\pm{0.2}$ & $145.7\pm{4.1}$ & $\underline{146.2}\pm{\underline{1.8}}$ & $116.2\pm7.0$ & $115.1\pm6.3$ & $133.8\pm5.1$ \vspace{3pt} \\
        ResNet+canonical & Equi. (can.) & $\underline{63.6}\pm\underline{0.1}$ & $\textbf{63.6}\pm\textbf{0.1}$ & $\underline{142.5}\pm\underline{2.0}$ & $\textbf{142.5}\pm\textbf{2.0}$ & $108.6\pm0.8$ & $108.6\pm0.8$ & $122.3\pm0.7$ \\
        VN-Transformer~\cite{assadd23_vntransformer} & Equi. (arch.) & $76.1\pm{0.9}$ & $76.1\pm{0.9}$ & $169.9\pm{2.9}$ & $169.9\pm{2.9}$ & $170.8\pm8.8$ & $170.8\pm8.8$ & $171.2\pm12.2$\\
        GotenNet~\cite{aykent25_gotennet} & Equi. (arch.) & $75.9\pm{0.6}$ & $75.9\pm{0.6}$ & $165.5\pm{1.0}$ & $165.5\pm{1.0}$ & $178.6\pm4.0$ & $178.6\pm4.0$ & $160.1\pm4.7$ \vspace{3pt}\\
        VN-Transformer & Hybrid & ${64.2}\pm{{0.1}}$ & $176.8\pm{1.0}$ & $153.2\pm{2.1}$ & $212.0\pm{2.2}$ & $137.9\pm10.7$ & $212.7\pm11.1$ & $165.7\pm6.4$ \\
        VN-Transformer+aug & Hybrid & $66.0\pm{0.4}$ & $66.0\pm{0.3}$ & $160.1\pm{7.4}$ & $160.2\pm{6.9}$ & $132.8\pm14.7$ & $133.6\pm14.7$ & $146.4\pm5.3$ \\
        GotenNet & Hybrid & $64.3\pm{0.3}$ & $177.5\pm{0.8}$ & ${143.7}\pm{{4.3}}$ & $206.0\pm{3.1}$ & $102.9\pm0.7$ & $182.0\pm2.9$ & $144.5\pm1.8$ \\
        GotenNet+aug & Hybrid & $65.8\pm{0.2}$ & $65.9\pm{0.3}$ & $151.1\pm{0.6}$ & ${150.8}\pm{{1.2}}$ & $\textbf{99.8}\pm\textbf{0.5}$ & $\textbf{99.6}\pm\textbf{0.5}$ & $\textbf{111.3}\pm\textbf{1.1}$ \\
        \midrule
        \multirow{2}{*}{FMPose*} & Vanilla (det.) & $59.1\pm0.1$ & $216.2\pm2.8$ & $134.3\pm0.2$ & $234.8\pm1.4$ & $106.6\pm0.8$ & $213.2\pm1.8$ & $156.3\pm1.4$ \\
         & Vanilla (prob.) & $\textbf{41.7}\pm\textbf{0.2}$ & $174.1\pm4.3$ & $106.4\pm0.2$ & $196.6\pm5.2$ & $78.6\pm0.9$ & $175.1\pm2.6$ & $128.6\pm1.7$ \\
        \multirow{2}{*}{FMPose*+aug} & Vanilla (det.) & $59.8\pm0.1$ & $60.0\pm0.1$ & $131.2\pm0.1$ & $131.0\pm0.2$ & $103.4\pm1.7$ & $102.9\pm0.6$ & $132.9\pm3.2$ \\
         & Vanilla (prob.) & $\underline{42.2}\pm\underline{0.2}$ & $\textbf{42.4}\pm\textbf{0.0}$ & $\textbf{100.3}\pm\textbf{0.5}$ & $\textbf{100.3}\pm\textbf{0.2}$ & $\textbf{72.3}\pm\textbf{1.2}$ & $\textbf{71.8}\pm\textbf{0.6}$ & $\underline{96.1}\pm\underline{1.8}$ \\
        \multirow{2}{*}{FMPose*+canonical} & Equi. (det.) & $61.2\pm0.1$ & $61.2\pm0.1$ & $130.7\pm0.6$ & $130.7\pm0.6$ & $103.3\pm0.5$ & $103.3\pm0.5$ & $124.5\pm0.9$ \\
         & Equi. (prob.) & $43.1\pm0.2$ & $\underline{43.1}\pm\underline{0.2}$ & $\underline{104.0}\pm\underline{0.5}$ & $\underline{104.0}\pm\underline{0.5}$ & $\underline{75.9}\pm\underline{0.6}$ & $\underline{75.9}\pm\underline{0.6}$ & $\textbf{96.0}\pm\textbf{1.3}$ \\
        \bottomrule
    \end{tabularx}
    \label{tab:test_protocol1}
\end{table*}
\vspace{-3pt}
\subsection{Metrics}
\vspace{-4pt}
We use the mean per-joint position error (MPJPE) as the main evaluation metric.
The MPJPE measures the average Euclidean distance between the estimated 3D human poses $\hat{\mathbf{X}}\in\mathbb{R}^{N\times 3}$ and ground truth $\mathbf{X}\in\mathbb{R}^{N\times 3}$ in millimetres (\unit{\milli\metre}):
\begin{equation}
    \textup{MPJPE}(\mathbf{X}, \hat{\mathbf{X}}) = \frac{1}{N} \sum_{n=1}^{N} \left\|\mathbf{X}_n-\hat{\mathbf{X}}_n\right\|_{2}~.
    \label{eq:mpjpe}
\end{equation}
We refer to the MPJPE metric as \textit{Protocol 1} (Table~\ref{tab:test_protocol1}).
In Supplementary~Section~\ref{supp:additional_quant}, we also report the Procrustes-aligned mean per-joint position error (PA-MPJPE), commonly known as \textit{Protocol 2}.
The PA-MPJPE measures prediction error after optimal similarity alignment (rotation, translation, and scale) to the ground truth.

\vspace{-2pt}
\subsection{Implementation Details}
\vspace{-2pt}
\noindent\textbf{Training details}\quad
We use PyTorch~\cite{paszke2019pytorch} in our experiments.
For Human3.6M and MPI-INF-3DHP, we use 2D poses detected by HRNet~\cite{cheng20_hrnet}, trained on the Human3.6M training set.
For EMDB, we instead use ground-truth 2D keypoints to isolate lifting performance from detection errors, since the trained HRNet performs poorly on the selected sequences containing natural full-body rotations.
For completeness, we also present the results using MPI-INF-3DHP ground-truth 2D keypoints as model inputs.
The input 2D poses are mean-centred and scaled by a single standard deviation computed over both coordinate dimensions, followed by pelvis centring (see Section~\ref{sec:geometrically_consistent} and Supplementary~ Section~\ref{supp:normalisation} for details).
The ground-truth 3D poses are expressed in camera coordinates and root-aligned at the origin.
The lifting models are trained to minimise the mean squared error (MSE) between the ground truth and estimated human poses, for a total of $100$ epochs on the training set of the Human3.6M dataset, with a batch size of $1024$ (reduced to $512$ for GotenNet due to computational constraints), Adam optimiser, an initial learning rate of $10^{-3}$, and the ExponentialLR scheduler with $\gamma = 0.96$.

For FMPose, a state-of-the-art reference method, we follow the official training protocol~\cite{le2026flow}, while applying the same input normalisation as for the other models. 
When ground-truth 2D keypoints are used for FMPose, we replicate each input pose to match the number of 2D samples expected by the model, rather than sampling from detector heatmaps.
For FMPose+canonical, the canonical rotation is computed from the argmax 2D pose and applied to all 2D input poses, with the inverse rotation applied to each predicted 3D hypothesis (see Supplementary~Section~\ref{supp:canonicalisation} for details).
%

\noindent\textbf{Data augmentation}\quad
When $+$\textbf{aug} is specified, we augment the training data by rotating the input 2D pose $X\in \mathbb{R}^{N\times 2}$ and the $xy$-coordinates of the corresponding ground-truth pose $\mathbf{X}\in \mathbb{R}^{N\times 3}$ by the same angle $\theta \sim \mathcal{U}[0, 2\pi)$.
Importantly, the rotation-based augmentation is applied to the samples on the fly, without increasing the total number of training samples per epoch, or explicitly covering all possible rotations for each sample.
This type of augmentation is commonly used in equivariance comparisons of point cloud models~\cite{deng2021vector, assadd23_vntransformer}.
For the \emph{Rotated} test sets, we independently sample rotation angles from the same distribution and apply each rotation jointly to the \emph{Original} 2D pose and its corresponding 3D ground truth.
Thus, \emph{Rotated} pose evaluation measures the rotation robustness of the lifting models, independently of the detector response to image rotations.
%

\noindent\textbf{Computational resources}\quad
All experiments are conducted on NVIDIA A100 GPUs (40 GB).
Each model is trained and evaluated with three different random seeds ($41$-$43$).
Table~\ref{tab:time} presents the training and inference times.
\subsection{Results and Discussion}
\label{sec:results}
\vspace{-4pt}
In this section, we analyse the effects of exact equivariance and rotation augmentation on prediction accuracy, robustness to rotations, cross-dataset generalisation, and computational efficiency.
The results under \textit{Protocol~1} are presented in Table~\ref{tab:test_protocol1} (mean and std across three runs), with \textit{Protocol~2} results provided in the Supplementary~Section~\ref{supp:additional_quant}.
The supplementary \textit{Protocol~2} results broadly support our observations, although individual model rankings differ.

\noindent\textbf{Rotation augmentation provides strong rotational robustness with only small changes in original-pose accuracy}\quad
Without rotation augmentation, typical lifting models exhibit substantial performance degradation under in-plane rotations.
On Human3.6M, MPJPE increases from $62.9$ to $209.6$\,mm for ResNet and from $65.9$ to $232.4$\,mm for PoseFormer (see Table~\ref{tab:test_protocol1}).
The non-augmented hybrid architectures exhibit similar degradation, with MPJPE increasing from $64.2$ to $176.8$\,mm for VN-Transformer hybrid and from $64.3$ to $177.5$\,mm for GotenNet hybrid.
A similar pattern is observed for FMPose, whose probabilistic variant increases from $41.7$ to $174.1$\,mm.

Rotation augmentation substantially improves accuracy on the \emph{Rotated} test sets.
On Human3.6M, it reduces MPJPE from $209.6$ to $64.3$\,mm for ResNet and from $232.4$ to $65.0$\,mm for PoseFormer, corresponding to reductions of approximately $69\%$ and $72\%$, respectively.
The hybrid models exhibit comparable improvements: augmentation reduces MPJPE on rotated Human3.6M poses from $176.8$ to $66.0$\,mm for VN-Transformer hybrid and from $177.5$ to $65.9$\,mm for GotenNet hybrid.
Both FMPose variants also benefit substantially; in particular, augmentation reduces the rotated-pose MPJPE of probabilistic FMPose from $174.1$ to $42.4$\,mm.
The resulting errors are close to those on the \emph{Original} test set, indicating that augmentation largely eliminates the loss of prediction accuracy under rotations.
Notably, the hybrid models achieve substantially lower MPJPE than their fully equivariant counterparts on \emph{Original} poses, but do not retain rotational robustness without augmentation.

Importantly, improved rotational robustness does not require a substantial sacrifice in accuracy on the original Human3.6M test distribution: augmentation increases the MPJPE of ResNet from $62.9$ to $64.2$\,mm, while slightly improving PoseFormer accuracy from $65.9$ to $65.1$\,mm.
Similarly, augmentation increases MPJPE on the \emph{Original} set by $1.8$ and $1.5$\,mm for VN-Transformer hybrid and GotenNet hybrid, respectively, and by $0.7$ and $0.5$\,mm for deterministic and probabilistic FMPose.
\begin{table}[t]
    \centering
    \footnotesize
    \vspace{-15pt}
    \caption{
   Training time per epoch and inference time per sample for the evaluated lifting models.  
   $^*$The fully equivariant GotenNet uses batch size 512 due to computational constraints.
   The best results are highlighted in \textbf{bold}, and the second-best results are \underline{underlined}.}
    \begin{tabularx}{\linewidth}{@{\extracolsep{\fill}}lccc}
        \toprule
        Method & Type & {Training} & {Inference} \\
        \midrule
        ResNet~\cite{martinez17_3dbaseline} & Vanilla & \textbf{11}\,s & \textbf{0.39}\,ms \\ 
        ResNet+aug & Vanilla &  34\,s & \textbf{0.39}\,ms \\
        PoseFormer~\cite{zheng21_poseformer} & Vanilla & {26}\,s & {2.23}\,ms \\ 
        PoseFormer+aug & Vanilla & 48\,s & \;{2.23}\,ms \vspace{3pt} \\ 
        ResNet+canonical & Equivariant (can.) & \underline{15}\,s & \underline{0.92}\,ms \\
        VN-Transformer~\cite{assadd23_vntransformer} & Equivariant (arch.) & 5\,m\;18\,s & 4.28\,ms \\ 
        GotenNet$^*$~\cite{aykent25_gotennet} & Equivariant (arch.) & 11\,m\;5\,s & 14.38\,ms \vspace{3pt}\\ 
        VN-Transformer & Hybrid & 46\,s & 4.68\,ms \\ 
        VN-Transformer+aug & Hybrid & 47\,s & 4.68\,ms \\ 
        GotenNet & Hybrid & 3\,m\;41\,s & 14.68\,ms \\ 
        GotenNet+aug & Hybrid & 3\,m\;43\,s & 14.68\,ms \\ 
        \midrule
        \multirow{2}{*}{FMPose} & Vanilla (det.) & 1\,m\;4\,s & 48.03\,ms\\
         & Vanilla (prob.) & 1\,m\;4\,s & 57.88\,ms\\
        \multirow{2}{*}{FMPose+aug} & Vanilla (det.) & 1\,m\;7\,s & 48.22\,ms\\
         & Vanilla (prob.) & 1\,m\;7\,s & 57.85\,ms \\
        \multirow{2}{*}{FMPose+canonical} & Equivariant (det.) & 1\,m\;22\,s & 48.48\,ms \\
         & Equivariant (prob.) & 1\,m\;22\,s & 59.99\,ms \\
        \bottomrule
    \end{tabularx}
    \vspace{-11pt}
    \label{tab:time}
\end{table}

\noindent\textbf{Exact equivariance: competitive accuracy is achievable via canonicalisation}\quad
On Human3.6M, the fully equivariant VN-Transformer and GotenNet result in MPJPEs of $76.1$ and $75.9$\,mm, respectively, on both \emph{Original} and \emph{Rotated} poses (see Table~\ref{tab:test_protocol1}).
Exact equivariance preserves MPJPE under the controlled rotation protocol, but does not guarantee low prediction error.
These architecture-based equivariant models are less accurate than the augmented vanilla models.
However, exact equivariance does not necessarily imply lower prediction accuracy: canonicalisation-based ResNet achieves $63.6$\,mm on both test sets, compared with $64.2$ and $64.3$\,mm for ResNet+aug.
For deterministic and probabilistic FMPose, canonicalisation yields $61.2$ and $43.1$\,mm on Human3.6M, respectively, compared with $59.8$ and $42.2$\,mm for augmentation on \emph{Original} poses.
These same-backbone comparisons show that exact equivariance need not entail a substantial accuracy degradation, and that the observed performance depends on the mechanism used to enforce it, as well as the backbone.

\noindent\textbf{Cross-dataset generalisation under controlled rotations depends on the equivariance mechanism and 2D input source}\quad
The MPI-INF-3DHP results provide a complementary evaluation of rotational robustness outside Human3.6M, using both HRNet detections and ground-truth 2D keypoints.
As on Human3.6M, rotation augmentation substantially reduces the degradation observed for vanilla models on the \emph{Rotated} test set.
However, the relative performance of augmentation and canonicalisation differs across backbones and 2D input sources.
With HRNet detections, ResNet+canonical achieves $142.5$\,mm on both \emph{Original} and \emph{Rotated} poses, compared with $152.3$ and $152.6$\,mm for ResNet+aug.
With ground-truth 2D inputs, their performance is nearly identical ($108.6$\,mm for ResNet+canonical and $108.5$/$107.2$\,mm for ResNet+aug).

Thus, the advantage of canonicalisation over augmentation for ResNet with HRNet detections largely disappears when ground-truth 2D keypoints are used.
For probabilistic FMPose, augmentation achieves lower MPJPE than canonicalisation with both HRNet detections ($100.3$ versus $104.0$\,mm) and ground-truth 2D inputs ($72.3$ versus $75.9$\,mm) on \emph{Original} poses.
Notably, GotenNet hybrid+aug achieves the lowest MPJPE among single-2D-input models with ground-truth 2D keypoints ($99.8$ and $99.6$\,mm).
Its non-augmented counterpart already achieves a low MPJPE on \emph{Original} poses ($102.9$\,mm), but exhibits substantial degradation under rotations ($182.0$\,mm), which augmentation largely eliminates.
\begin{figure}[t]
    \centering
    \includegraphics[width=\linewidth]{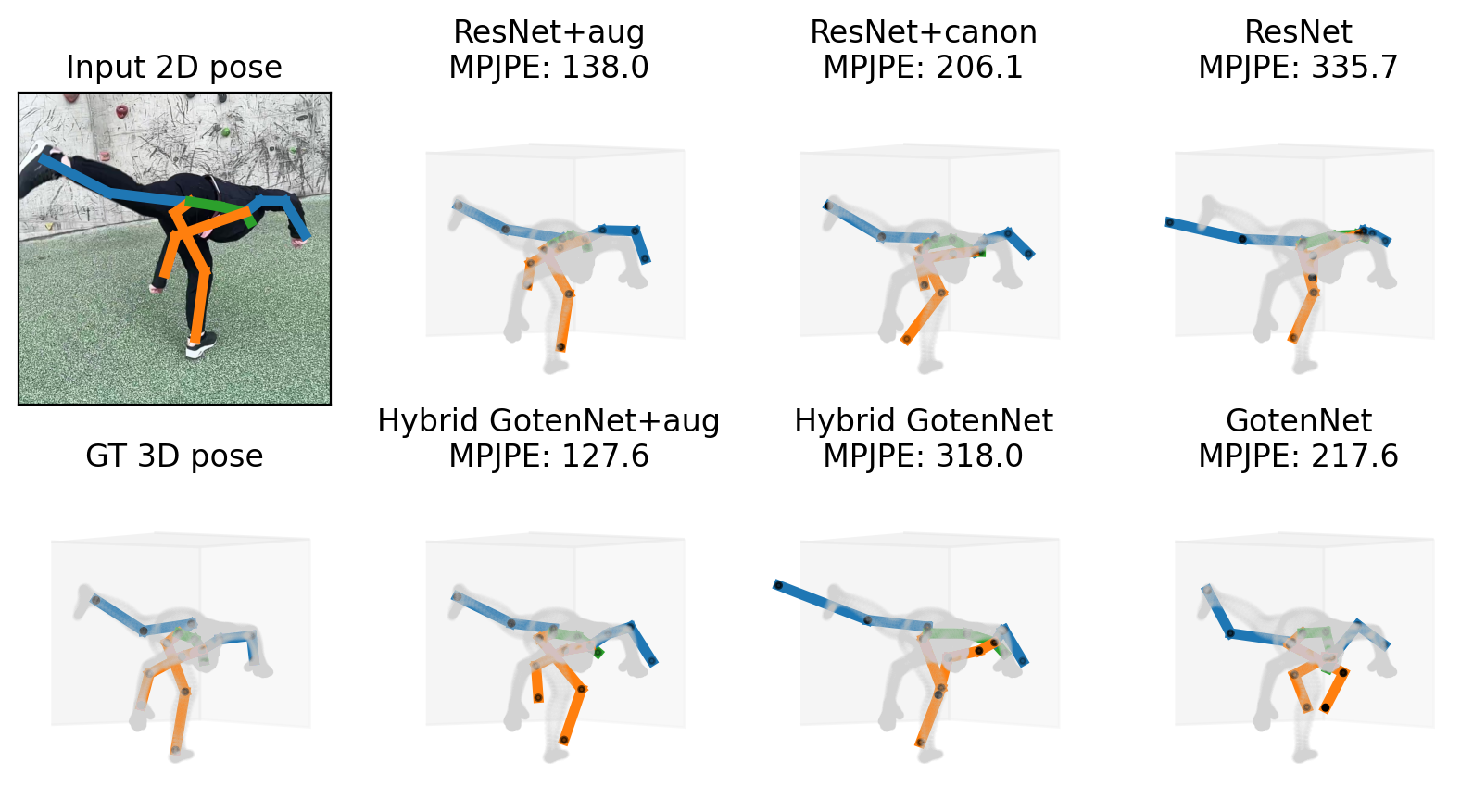}
    \vspace{+4pt}    \includegraphics[width=\linewidth]{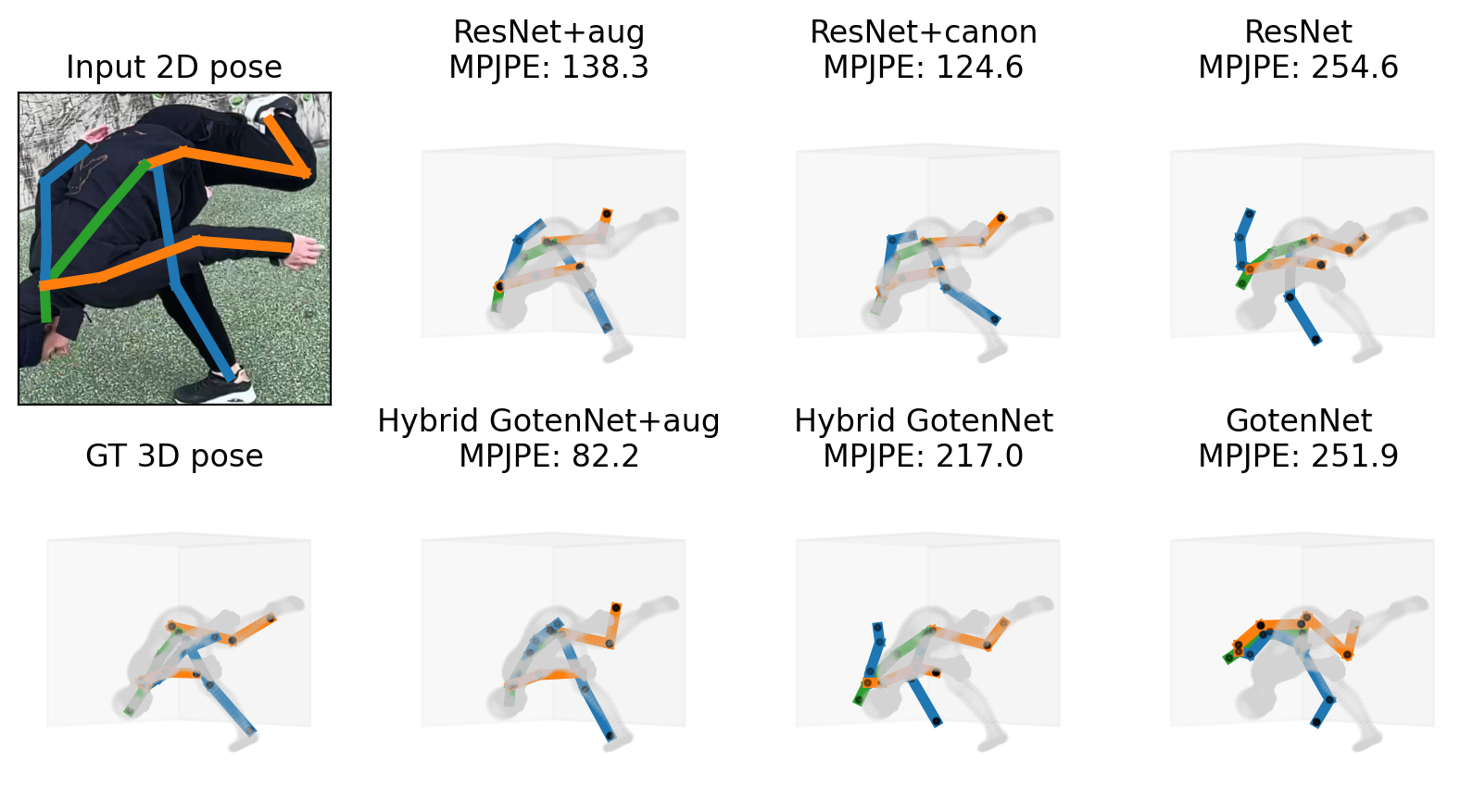}
    \vspace{-25pt}
    \caption{
    Examples of challenging poses with \textit{natural} rotations from the EMDB \texttt{69\_outdoor\_cartwheel} sequence.
    Comparison of 3D pose predictions from the evaluated models against ground truth, shown separately and overlaid in grey.
    }
    \vspace{-12pt}
    \label{fig:qual_emdb}
\end{figure}
\\\noindent\textbf{Natural full-body rotations: augmentation and canonicalisation improve accuracy}\quad
We additionally evaluate the selected EMDB sequences containing natural full-body rotations, using ground-truth 2D keypoints to evaluate lifting accuracy without errors from 2D pose detection (see the last column in Table~\ref{tab:test_protocol1}).
Compared with vanilla ResNet, both augmentation and canonicalisation reduce MPJPE, from $155.3$\,mm to $114.3$ and $122.3$\,mm, respectively.
Notably, GotenNet hybrid+aug achieves the lowest MPJPE among the single-2D-input models ($111.3$\,mm), improving upon its non-augmented counterpart ($144.5$\,mm).
For probabilistic FMPose, the corresponding errors decrease from $128.6$\,mm to $96.1$ and $96.0$\,mm.

These results show that augmentation improves accuracy on the selected naturally rotated sequences across vanilla, hybrid, and probabilistic models, while canonicalisation also improves accuracy within the evaluated ResNet and FMPose backbones.
Broader conclusions about real-world generalisation require evaluation on more such data with 3D ground truth.
Qualitative examples are shown in Figures~\ref{fig:teaser}~and~\ref{fig:qual_emdb}, with additional examples, including SportsCap, in Supplementary~Section~\ref{supp:additional_qual}.
\\\noindent\textbf{Augmentation and canonicalisation for computationally efficient rotation robustness}\quad
Table~\ref{tab:time} illustrates the computational differences between the evaluated methods.
For ResNet, rotation augmentation increases training time from $11$ to $34$\,s per epoch, without increasing the measured inference time ($0.39$\,ms per sample).
Canonicalisation, in contrast, requires $15$\,s per epoch and introduces an inference overhead ($0.92$\,ms per sample), while providing exact equivariance.
The fully equivariant VN-Transformer and GotenNet require $4.28$ and $14.38$\,ms per sample, respectively, illustrating the higher inference costs of the evaluated architecture-based equivariant models.
The hybrid models also incur higher inference costs than the vanilla ResNet and PoseFormer baselines.
FMPose has a considerably higher inference cost, particularly in its probabilistic setting, with canonicalisation introducing a small additional overhead.

\begin{figure}[t]
    \centering
        \begin{subfigure}[t]{\linewidth}
        \centering
        \includegraphics[width=\linewidth]{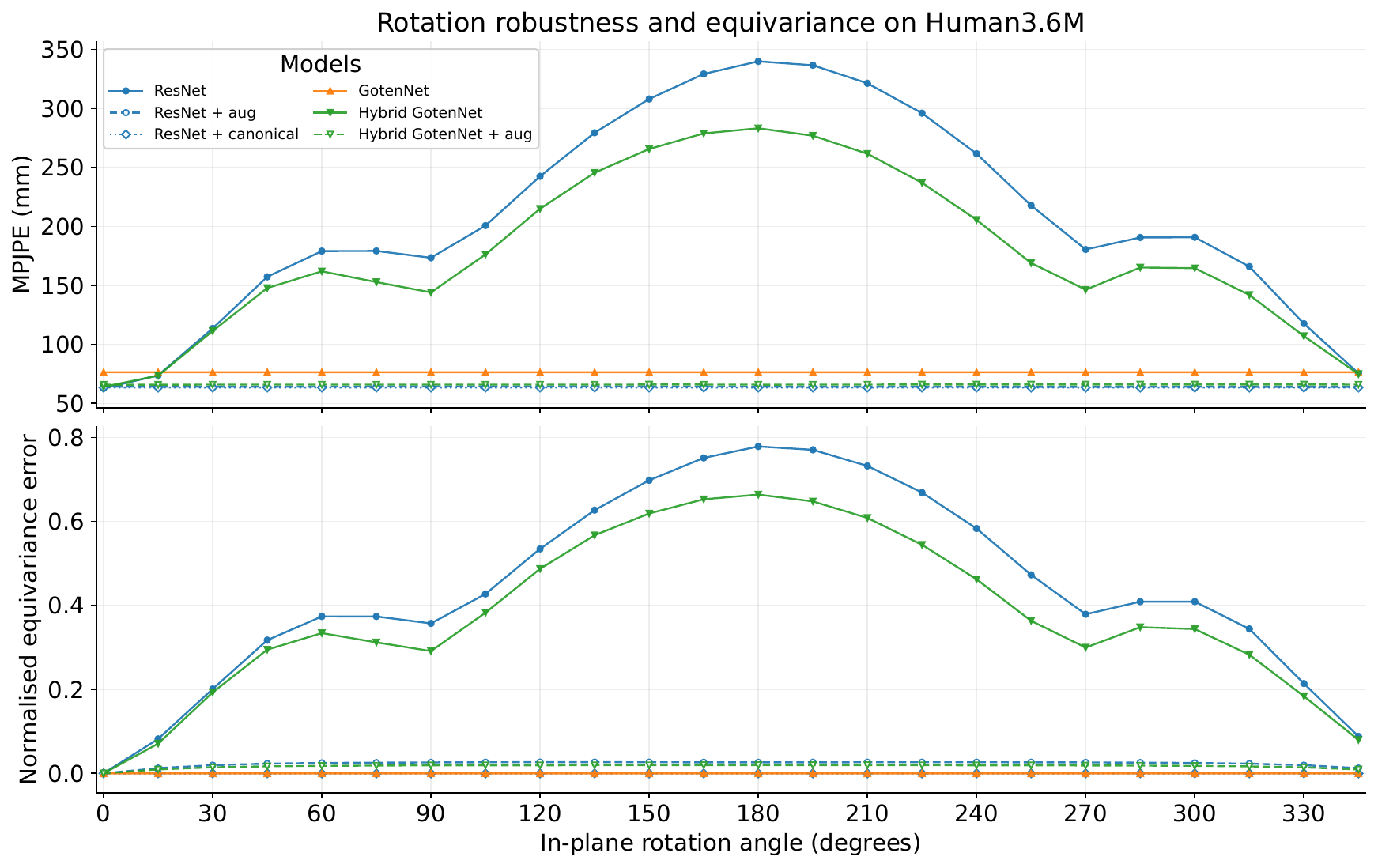}
        \caption{
  Rotation robustness analysis. Top: MPJPE as a function of rotation angle. Bottom: corresponding normalised equivariance error.
        }
        \label{fig:rotation_curves}
    \end{subfigure}
    \hfill
    \\
        \begin{subfigure}[t]{\linewidth}
        \centering
        \includegraphics[width=\linewidth]{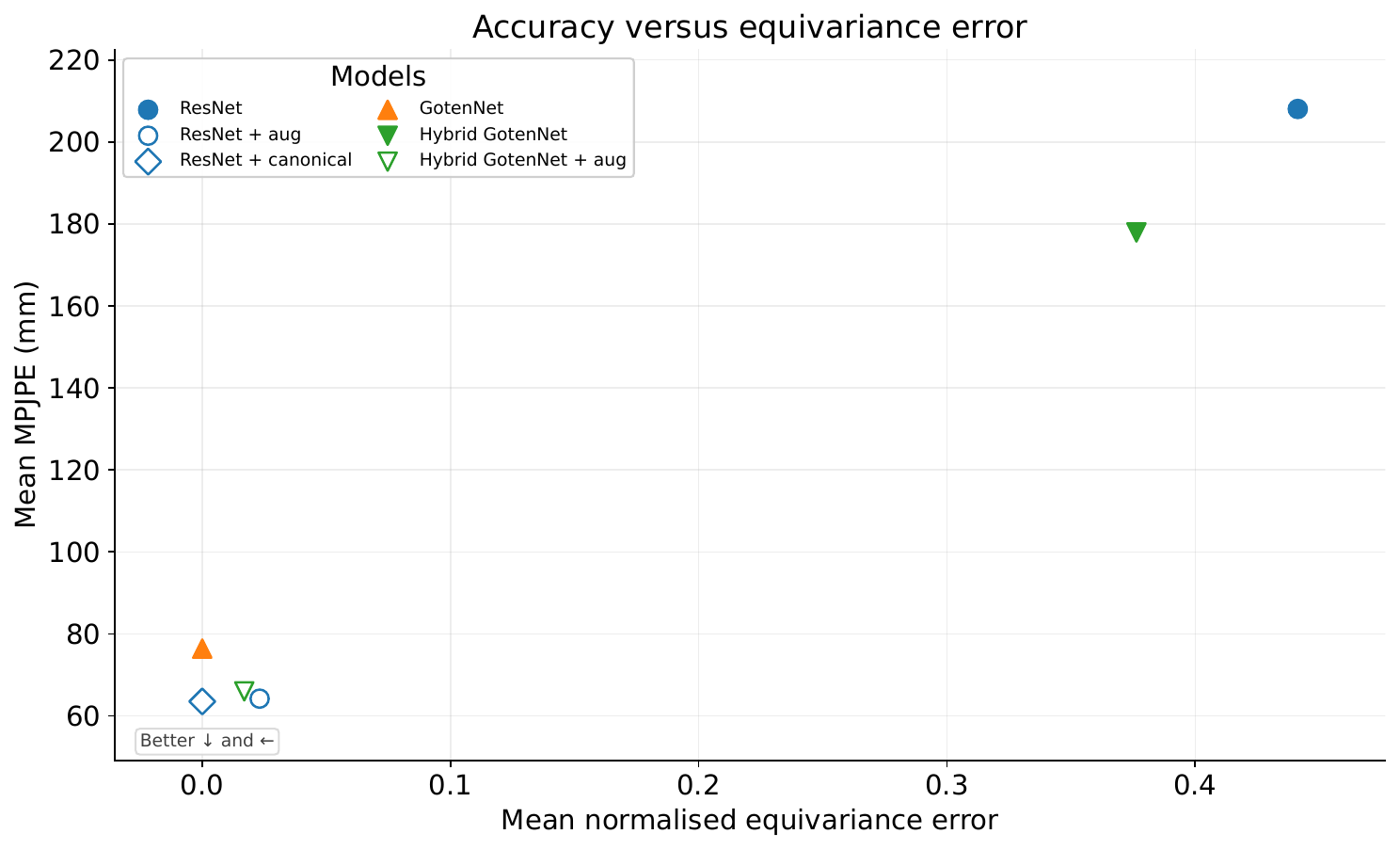}
        \caption{
        Accuracy vs. equivariance error. Each point represents a model, showing mean MPJPE (lower is better) versus mean normalised equivariance error (lower is better), averaged over in-plane rotations.
        }
        \vspace{-5pt}
        \label{fig:tradeoff}
    \end{subfigure}
    \caption{
    Rotation equivariance diagnostics on Human3.6M.
    We rotate each input 2D pose by 24 angles $\theta \in [0^\circ, 360^\circ)$ with a fixed step and evaluate models on the test set.
    }
    \vspace{-15pt}
    \label{fig:rotation_diagnostics}
\end{figure}

\noindent\textbf{Rotation equivariance diagnostics}\quad
To distinguish rotational robustness from equivariance, we evaluate the lifting models under $24$ uniformly spaced in-plane rotations $\theta \in [0, 2\pi)$ using the Human3.6M test set.
For each 2D pose $X \in \mathbb{R}^{N \times 2}$ and lifting model $f: \mathbb{R}^{N\times2} \rightarrow \mathbb{R}^{N\times3}$, we measure MPJPE (Eq.~\eqref{eq:mpjpe}) and the normalised equivariance error
\begin{equation}
    \label{eq:equivariance_error}
    \epsilon_\text{eq}(X, \theta) = \frac{\sum_{n=1}^N \left\| f(X\,R^\top_\theta)_n -
    \left( f(X) \begin{bmatrix}
        R^\top_\theta & 0 \\
        0 & 1
    \end{bmatrix} \right)_n
    \right\|_{2}}
    {\sum_{n=1}^N \left\| f(X)_n \right\|_{2}}~.
\end{equation}
Figure~\ref{fig:rotation_diagnostics} shows MPJPE and equivariance error across rotation angles.
Without augmentation, ResNet and Hybrid GotenNet exhibit substantial variation in MPJPE and large equivariance errors.
Augmentation substantially reduces both, yielding nearly constant MPJPE and low but nonzero equivariance errors, indicating approximate rather than exact equivariance.
In contrast, both architecture-based GotenNet and canonicalisation-based ResNet achieve zero equivariance error, up to numerical precision.
Despite this shared property, ResNet+canonical achieves MPJPE comparable to ResNet+aug, whereas GotenNet has higher prediction error.
The accuracy-vs-equivariance error comparison further shows that exact equivariance can coexist with competitive prediction accuracy, while rotational robustness alone does not establish exact equivariance.

\noindent\textbf{Ablation studies}\quad
Additional ablations of input depth initialisation for architecture-based equivariant models and equivariant feature selection in hybrid models
are provided in Supplementary~Section~\ref{supp:ablations}.
Neither alternative yields consistent improvements over the configurations used in the main comparisons.
\vspace{-8pt}
\section{Conclusions}
\label{sec:conclusions}
\vspace{-5pt}
In this work, we show that representative 2D-to-3D lifting models, including a state-of-the-art probabilistic method, exhibit substantial performance degradation under in-plane rotations, and investigate rotation equivariance as an inductive bias for improving robustness.
We systematically compare rotation augmentation, architecture-based equivariance, and canonicalisation across lifting models and datasets.
Our results show that augmentation provides strong rotational robustness with little loss of original-pose accuracy, while canonicalisation achieves exact equivariance with competitive prediction accuracy.
Both approaches improve performance for probabilistic lifting and real-world sequences containing natural full-body rotations.
Thus, exact equivariance does not inherently impose an accuracy penalty, although its effectiveness depends on how it is enforced.
Direct diagnostics further demonstrate that rotational robustness alone does not establish exact equivariance.
Future work could extend this analysis to temporal and end-to-end lifting models, additional input representations, and broader evaluations of natural rotations.

\noindent\textbf{Limitations}\quad
Our quantitative evaluation of natural full-body rotations is constrained by the limited availability of suitable sequences with 3D ground truth.
We therefore evaluate two selected EMDB sequences and provide additional qualitative examples from SportsCap, for which 3D ground truth is unavailable.
Our architectural comparisons cover representative lifting models and equivariant designs.


\section*{Acknowledgments}
This work was supported by the Wallenberg AI, Autonomous Systems and Software Program (WASP), by the Swedish Research Council (VR) through a grant for the projects Uncertainty-Aware Transformers for Regression Tasks in Computer Vision (2022-04266) and Dorsal Stream Robot Vision (2022-04206), and the strategic research environment ELLIIT. 
The computations were enabled by resources provided by the National Academic Infrastructure for Supercomputing in Sweden (NAISS), partially funded by the Swedish Research Council through grant agreement no. 2022-06725.

{
    \bibliographystyle{ieeenat_fullname}
    \bibliography{main}
}

\clearpage
\setcounter{page}{1}
\maketitlesupplementary
\section{Input pose normalisation}
\label{supp:normalisation}
We describe the normalisation of 2D keypoints used as model inputs and show that it preserves the SO(2) action defined in Eq.~\eqref{eq:geometrical_consistency}. 
Let $P\in\mathbb{R}^{N\times2}$ denote a detected 2D pose, with $N=16$ joints, and let $P_\text{pelvis}$ denote the pelvis joint. 

We first subtract the mean joint position,
\begin{equation}
   \widetilde P=P-\mathbf{1}\mu(P)~, \quad \text{where} \quad \mu(P)=\frac{1}{N}\sum_{n=1}^{N}P_n~, \qquad 
\label{eq:centering}
\end{equation}
with $\mathbf{1}\in\mathbb{R}^{N\times1}$ denoting a vector of ones. 
We then compute a single standard deviation across both coordinate axes and all joints:
\begin{equation}
s(P)= \sqrt{\frac{1}{2N} \sum_{n=1}^{N}\lVert\widetilde P_n\rVert_2^2}~.
\end{equation}
For nondegenerate poses with $s(P)>0$, the normalised input pose is obtained by dividing by this scalar and subsequently centring at the pelvis:
\begin{equation}
X=\mathcal{N}(P) = \frac{\widetilde P}{s(P)}  -\mathbf{1}\frac{\widetilde P_\text{pelvis}}{s(P)} = \frac{P-\mathbf{1}P_\text{pelvis}}{s(P)}~.
\end{equation}
Unlike independent standardisation of the horizontal and vertical coordinates, this normalisation uses the same scale factor for both axes.

\subsection{SO(2) consistency} 
For any $R_\theta\in \text{SO}(2)$, the mean joint position transforms as
\begin{equation}
    \mu(PR_\theta^\top)=\mu(P)R_\theta^\top~.
\end{equation}
Consequently, the centred pose in Eq.~\eqref{eq:centering} satisfies
\begin{equation}
\widetilde{PR_\theta^\top} = \widetilde P R_\theta^\top~.
\end{equation}
Since rotations preserve Euclidean norms, 
\begin{equation}
s(PR_\theta^\top)^2 = 
\frac{1}{2N} \sum_{n=1}^{N} \lVert \widetilde P_nR_\theta^\top \rVert_2^2 = s(P)^2~.
\end{equation}

The pelvis position also transforms covariantly. 
Therefore, 
\begin{equation}
\mathcal N(PR_\theta^\top) = \mathcal N(P)R_\theta^\top~.
\end{equation}

Thus, the input normalisation commutes with global SO(2) rotations and does not introduce an axis-dependent distortion into the transformation considered in Eq.~\eqref{eq:geometrical_consistency}.

\subsection{Coordinate interpretation} 
The models take as input the normalised, pelvis-centred 2D poses $X$, rather than raw image coordinates. 
Accordingly, the SO(2) transformations used for augmentation and evaluation act about the pelvis in normalised pose space, independently of the original image-coordinate origin. 
This establishes the mathematical consistency of the preprocessing with Eq.~\eqref{eq:geometrical_consistency}, without requiring rotations about an image or crop centre to correspond to physical camera rotations.

\section{Canonicalisation}
\label{supp:canonicalisation}
In this section, we describe the parameter-free canonicalisation used to construct SO(2)-equivariant models from the vanilla lifting backbones, and establish its equivariance under the action defined in Eq.~\eqref{eq:geometrical_consistency}.

\subsection{Canonical rotation}

Let $X\in\mathbb{R}^{N\times 2}$ denote a normalised,
pelvis-centred 2D pose. We define the anatomical reference vectors
\begin{equation}
\begin{aligned}
v_1(X) &= X_{\mathrm{neck}}-X_{\mathrm{pelvis}}~,\\
v_2(X) &=
\frac{X_{\mathrm{lshoulder}}+X_{\mathrm{rshoulder}}}{2}
-
\frac{X_{\mathrm{lhip}}+X_{\mathrm{rhip}}}{2}~.
\end{aligned}
\label{eq:canonical_references}
\end{equation}
We select $v(X)=v_1(X)$ when $\|v_1(X)\|_2>0$ and
$v(X)=v_2(X)$ otherwise. The construction below is defined
on the domain
\begin{equation}
\mathcal{D}=
\left\{
X:\|v_1(X)\|_2>0
\ \text{or}\
\|v_2(X)\|_2>0
\right\}.
\label{eq:canonical_domain}
\end{equation}

For $X\in\mathcal{D}$, let $v(X)=(v_x,v_y)$ and $r=\|v(X)\|_2$. 
We compute the rotation
\begin{equation}
R_C(X)= \frac{1}{r}
\begin{bmatrix}
v_y & v_x\\
-v_x & v_y
\end{bmatrix}
\in \text{SO}(2)~,
\label{eq:canonical_rotation}
\end{equation}
which satisfies
\begin{equation}
v(X)\,R_C(X)=(0,r).
\label{eq:canonical_alignment}
\end{equation}
Thus, the selected anatomical reference is aligned with the
positive vertical axis. The canonicalised input is
$X_C=XR_C(X)$.

\subsection{SO(2) equivariance}

We show that the canonicalisation-based predictor satisfies
Eq.~\eqref{eq:geometrical_consistency} for all $X\in\mathcal{D}$.

For any $R\in \text{SO}(2)$, both reference vectors transform
covariantly:
\begin{equation}
v_i(XR^\top)=v_i(X)R^\top,
\qquad i\in\{1,2\}.
\label{eq:reference_covariance}
\end{equation}
Since rotations preserve Euclidean norms, the reference-selection
rule is unchanged by the transformation. Consequently,
\begin{equation}
v(XR^\top)=v(X)R^\top.
\label{eq:selected_reference_covariance}
\end{equation}

The canonical rotation is the unique element of SO(2) that maps the selected nonzero reference vector to $(0,r)$.
By Eq.~\eqref{eq:selected_reference_covariance}, the matrix
$R\,R_C(X)$ maps $v(XR^\top)$ to the same target:
\begin{equation}
v(XR^\top)\,R\,R_C(X)
=
v(X)\,R_C(X)
=
(0,r).
\end{equation}
Uniqueness therefore gives
\begin{equation}
R_C(XR^\top)=R~R_C(X).
\label{eq:canonical_covariance}
\end{equation}
It follows that the canonicalised input is rotation-invariant:
\begin{equation}
\begin{aligned}
XR^\top ~ R_C(XR^\top)
&=XR^\top \, R\,R_C(X) =X\,R_C(X).
\end{aligned}
\label{eq:canonical_invariance}
\end{equation}

Let $f:\mathbb{R}^{N\times2}\rightarrow
\mathbb{R}^{N\times3}$ be an arbitrary vanilla lifting model,
and define
\begin{equation}
Q(R)=
\begin{bmatrix}
R&0\\
0&1
\end{bmatrix}.
\end{equation}
The canonicalisation-based predictor is
\begin{equation}
g(X)=f\bigl(XR_C(X)\bigr) \, Q(R_C(X))^\top.
\label{eq:canonical_predictor}
\end{equation}
Using Eqs.~\eqref{eq:canonical_covariance} and
\eqref{eq:canonical_invariance}, we obtain
\begin{equation}
\begin{aligned}
g(XR^\top)
&=
f\bigl(XR^\top R_C(XR^\top)\bigr) ~
Q(R_C(XR^\top))^\top\\
&=
f\bigl(XR_C(X)\bigr) ~
Q(RR_C(X))^\top\\
&=
f\bigl(XR_C(X)\bigr) ~
Q(R_C(X))^\top Q(R)^\top\\
&=g(X)~Q(R)^\top.
\end{aligned}
\label{eq:canonical_equivariance}
\end{equation}
Hence, $g$ satisfies Eq.~\eqref{eq:geometrical_consistency} for every $X\in\mathcal{D}$,
independently of the architecture or parameters of $f$.
In particular, the predicted $xy$-coordinates transform
equivariantly, while the predicted depth remains invariant.

For FMPose~\cite{le2026flow}+canonical, a single canonical rotation is computed from the argmax 2D pose $X_{\max}$ and applied to all 2D input poses $X_k$. 
Under a global in-plane rotation $R$, Eq.~\eqref{eq:canonical_covariance} gives $R_C(X_{\max}R^\top)=RR_C(X_{\max})$, so each canonicalised input remains unchanged:
\begin{equation}
(X_kR^\top)R_C(X_{\max}R^\top)=X_kR_C(X_{\max}).
\end{equation}
Thus, the predicted 3D hypotheses transform equivariantly after applying the inverse canonical rotation.
This holds for individual probabilistic hypotheses when inference randomness is shared between the original and rotated inputs, and more generally, the predicted distribution is SO(2)-equivariant.

\subsection{Implementation and degeneracies}
Canonicalisation introduces no learnable parameters.
It consists of selecting an anatomical reference, computing a two-dimensional rotation, applying it to the input pose, and applying the inverse rotation to the predicted $xy$-coordinates.
Our implementation uses the negative vertical direction as its canonical target, corresponding to the image-coordinate convention. 
This replaces $R_C(X)$ in Eq.~\eqref{eq:canonical_rotation} by $-R_C(X)$, which is also an element of SO(2). 
The equivariance derivation remains unchanged.

The guarantee applies whenever at least one reference vector is nonzero. 
If both references vanish, the canonical orientation is undefined, and the construction does not provide an equivariance guarantee.
In finite-precision arithmetic, the identities above hold up to numerical error.

To assess the coverage of the anatomical references $v_1$ and $v_2$, we report statistics for Human3.6M and MPI-INF-3DHP in Table~\ref{tab:canonical_reference_stats}. 
The Human3.6M diagnostics motivated our choice of the pelvis-to-neck vector as the primary reference and the vector from the hip midpoint to the shoulder midpoint as a fallback.
On Human3.6M, the primary reference vanishes for only 34 training and 12 test poses.
On MPI-INF-3DHP, it is nonzero for all 2,875 test poses using either HRNet-detected or ground-truth 2D keypoints. 
In all evaluated cases, at least one reference vector is nonzero, so no simultaneous degeneracies occur.
\begin{table}[t]
    \centering
    \caption{Reference-vector coverage for SO(2) canonicalisation on Human3.6M and MPI-INF-3DHP (test-set HRNet detected and ground-truth 2D keypoints).
    The pelvis-to-neck vector is selected whenever nonzero; otherwise, the vector from the hip midpoint to the shoulder midpoint is used.
    Degenerate cases occur when both reference vectors vanish (their norms $r=0$).}
    \small
    \resizebox{\columnwidth}{!}{%
    \begin{tabular}{lrrrr}
        \toprule
        & \multicolumn{2}{c}{Human3.6M}
        & \multicolumn{2}{c}{MPI-INF-3DHP} \\
        \cmidrule(lr){2-3}\cmidrule(lr){4-5}
        Reference selection
        & Train & Test & HRNet & GT 2D \\
        \midrule
        Total poses             & 1,559,752 & 543,344 & 2,875 & 2,875 \\
        Primary reference used  & 1,559,718 & 543,332 & 2,875 & 2,875 \\
        Fallback reference used & 34        & 12      & 0     & 0     \\
        Both references zero    & 0         & 0       & 0     & 0     \\
        \bottomrule
    \end{tabular}%
    }
    \label{tab:canonical_reference_stats}
\end{table}
\section{Additional quantitative results}
\vspace{-2pt}
\label{supp:additional_quant}
In Table~\ref{tab:test_protocol2}, we present the \textit{Protocol 2} results, \ie Procrustes-aligned mean per-joint position error (PA-MPJPE), which measures prediction error after optimal similarity alignment (rotation, translation, and scale) to the ground truth.
\begin{table*}[ht]
    \centering
    \footnotesize
     \caption{
    Quantitative results with evaluation \textit{Protocol~2} (PA-MPJPE, $\downarrow$) on the test sets of Human3.6M and MPI-INF-3DHP, as well as a subset of EMDB. 
    The best and second-best results among methods using a single 2D pose as input are highlighted in \textbf{bold} and \underline{underlined}, respectively.
    *FMPose is included as a reference state-of-the-art method and differs from the other models by operating on multiple 2D pose samples (from a single image) and producing multiple 3D hypotheses; we report single (deterministic) and oracle best-of-hypotheses (probabilistic); see Section~\ref{sec:vanilla} for details; the \textbf{best}  and \underline{second-best} results for FMPose are highlighted separately from the other models.
    $^\dagger$ Evaluations using ground-truth 2D keypoints instead of HRNet detections.
    }
    \vspace{-5pt}
    \begin{tabularx}{\linewidth}{@{\extracolsep{\fill}}llccccccc}
        \toprule
        \multirow{2}{*}{Method} & \multirow{2}{*}{Type} & \multicolumn{2}{c}{Human3.6M} & \multicolumn{2}{c}{MPI-INF-3DHP} & \multicolumn{2}{c}{MPI-INF-3DHP$^\dagger$} & EMDB$^\dagger$ \\
        & & \textit{Original} & \textit{Rotated} & \textit{Original} & \textit{Rotated} & \textit{Original} & \textit{Rotated} &  \textit{Original}\\
        \midrule
        ResNet~\cite{martinez17_3dbaseline} & Vanilla & $\textbf{44.5}\pm{\textbf{0.1}}$ & $109.1\pm{0.8}$ & ${99.3}\pm{0.4}$ & $134.6\pm{1.0}$ & $68.2 \pm 1.4$ & $114.6 \pm 0.9$ & $85.2\pm0.1$ \\
        ResNet+aug & Vanilla & $46.2\pm{0.0}$ & $\underline{46.2}\pm{\underline{0.1}}$ & $102.6\pm{2.0}$ & ${102.6}\pm{{1.4}}$ & $68.0 \pm 0.1$ & $\underline{67.1} \pm \underline{0.4}$
        & $\textbf{70.9}\pm\textbf{1.0}$ \\
        PoseFormer~\cite{zheng21_poseformer} & Vanilla & $47.9\pm{0.1}$ & $127.0\pm{1.6}$ & $\underline{97.4}\pm{\underline{1.0}}$ & $141.4\pm{1.0}$ & $78.6 \pm 1.0$ & $133.8 \pm 0.9$ & $98.0\pm2.1$ \\
        PoseFormer+aug & Vanilla & $46.6\pm{0.1}$ & ${46.5}\pm{{0.2}}$ & $101.3\pm{2.9}$ & $\underline{101.9}\pm{\underline{1.9}}$ & $76.5 \pm 4.6$ & $76.1 \pm 4.0$ & $86.2\pm3.9$ \vspace{3pt} \\
        ResNet+canonical & Equi. (can.) & $\underline{44.6}\pm\underline{0.1}$ & $\textbf{44.6}\pm\textbf{0.1}$ & $\textbf{96.9}\pm\textbf{1.5}$ & $\textbf{96.9}\pm\textbf{1.5}$ & $70.5 \pm 0.7$ & $70.5 \pm 0.7$ & $74.6\pm0.4$ \\
        VN-Transformer~\cite{assadd23_vntransformer} & Equi. (arch.) & $55.4\pm{0.5}$ & $55.4\pm{0.5}$ & $121.0\pm{2.4}$ & $121.0\pm{2.4}$ & $106.4 \pm 4.5$ & $106.4 \pm 4.5$ & $112.9\pm4.7$\\
        GotenNet~\cite{aykent25_gotennet} & Equi. (arch.) & $55.1\pm{0.5}$ & $55.1\pm{0.5}$ & $122.3\pm{0.6}$ & $122.3\pm{0.6}$ & $111.9 \pm 2.0$ & $111.9 \pm 2.0$ & $100.7\pm2.3$ \vspace{3pt}\\
        VN-Transformer & Hybrid & ${45.6}\pm{{0.1}}$ & $101.2\pm{0.7}$ & $106.2\pm{0.8}$ & $134.4\pm{1.5}$ & $70.0 \pm 2.3$ & $110.8 \pm 3.3$ 
        & $90.8\pm1.2$ \\
        VN-Transformer+aug & Hybrid & $46.8\pm{0.2}$ & $46.8\pm{0.2}$ & $110.6\pm{4.5}$ & $110.1\pm{3.6}$ & $69.1 \pm 0.6$ & $70.0 \pm 0.5$ & $82.3\pm0.5$ \\
        GotenNet & Hybrid & ${45.6}\pm{{0.1}}$ & $102.2\pm{1.0}$ & $101.3\pm{3.5}$ & $130.1\pm{1.8}$ & $\underline{67.3} \pm \underline{0.6}$ & $107.9 \pm 1.5$ & $79.6\pm1.1$\\
        GotenNet+aug & Hybrid & $46.9\pm{0.1}$ & $47.0\pm{0.1}$ & $105.2\pm{1.0}$ & $104.1\pm{1.1}$ & $\textbf{66.1} \pm \textbf{0.5}$ & $\textbf{65.6} \pm \textbf{0.1}$ & $\underline{71.2}\pm\underline{0.1}$ \\
        \midrule
        \multirow{2}{*}{FMPose*} & Vanilla (det.) & $41.5\pm0.1$ & $103.9\pm1.9$ & $84.0\pm0.3$ & $121.6\pm0.7$ & $65.3 \pm 0.4$ & $107.1 \pm 0.5$ & $82.5\pm1.3$ \\
         & Vanilla (prob.) & $\textbf{30.4}\pm\textbf{0.1}$ & $85.2\pm1.4$ & $67.0\pm0.1$ & $102.1\pm0.8$ & $48.7 \pm 0.1$ & $88.6 \pm 0.5$ & $64.8\pm0.8$ \\
        \multirow{2}{*}{FMPose*+aug} & Vanilla (det.) & $42.5\pm0.1$ & $42.4\pm0.1$ & $82.6\pm0.3$ & $82.4\pm0.2$ & $61.2 \pm 0.4$ & $61.1 \pm 0.4$ & $69.7\pm1.9$ \\
         &  Vanilla (prob.) & $31.4\pm0.1$ & $\underline{31.4}\pm\underline{0.0}$ & $\textbf{63.3}\pm\textbf{0.2}$ & $\textbf{63.3}\pm\textbf{0.1}$ & $\textbf{42.7} \pm \textbf{0.3}$ & $\textbf{42.6} \pm \textbf{0.3}$ 
         & $\textbf{51.0}\pm\textbf{1.4}$ \\
         \multirow{2}{*}{FMPose*+canonical} & Equi. (det.) & $42.5\pm0.5$ & $42.5\pm0.5$ & $82.4\pm0.1$ & $82.4\pm0.1$ & ${62.8}\pm {0.4}$ & ${62.8}\pm {0.4}$ & $69.2\pm0.2$ \\
         & Equi. (prob.) & $\underline{30.9}\pm\underline{0.1}$ & $\textbf{30.9}\pm\textbf{0.1}$ & $\underline{65.7}\pm\underline{0.2}$ & $\underline{65.7}\pm\underline{0.2}$ & $\underline{46.0} \pm \underline{0.3}$ & $\underline{46.0}\pm \underline{0.3}$ 
         & $\underline{53.2}\pm\underline{0.6}$ \\
        \bottomrule
    \end{tabularx}
   \vspace{-10pt}
    \label{tab:test_protocol2}
\end{table*}
\begin{table}[t]
    \centering
    \footnotesize
    \caption{
    Ablation study for the depth initialisation for fully equivariant models: Quantitative results with evaluation \textit{Protocol~1} on the \textit{Original} and \textit{Rotated} test sets of Human3.6M.
    \textit{Fixed rand init} $z$ means fixed initial depth, initialised randomly with the model instantiation, \textit{learn init} $z$ means learnable initial depth, \textit{homogeneous} means the initial 2D input is expressed in the homogeneous coordinates, \ie with $z=1$ for all the joints (N/A for GotenNet due to the joint-wise differences used in the model, which leads to the third coordinate in the output always being zero), and the selected approach $z=\sqrt{x^2 + y^2}$ uses the Euclidean norm of the 2D coordinates of each joint as the input depth.
    }
    \vspace{-5pt}
    \begin{tabularx}{\linewidth}{@{\extracolsep{\fill}}ll cc}
        \toprule
        \multirow{2}{*}{Method} & \multirow{2}{*}{Depth initialisation} & \multicolumn{2}{c}{Human3.6M MPJPE ($\downarrow$)} 
        \\
        & & \textit{Original} & \textit{Rotated} \\
        \midrule
        VN-Transformer & Homogeneous & $77.0\pm{0.1}$ & $77.0\pm{0.1}$ \\
        VN-Transformer & Fixed rand init $z$ & ${76.4}\pm{{1.2}}$ & ${76.4}\pm{{1.2}}$ \\
        VN-Transformer & Learn init $z$& ${76.4}\pm{{0.4}}$ & ${76.4}\pm{{0.4}}$ \\ 
        VN-Transformer & $z=\sqrt{x^2 + y^2}$& ${76.1}\pm{{0.9}}$ & ${76.1}\pm{{0.9}}$ \\
        \midrule
        GotenNet & Fixed rand init $z$ & $76.7\pm{1.3}$ & $76.7\pm{1.3}$ \\
        GotenNet & Learn init $z$ & $77.2\pm0.5$ & $77.2\pm0.5$ \\
         GotenNet & $z=\sqrt{x^2 + y^2}$& ${75.9}\pm{{0.6}}$ & ${75.9}\pm{{0.6}}$ \\ 
        \bottomrule
    \end{tabularx}
    \vspace{-15pt}
    \label{tab:equi_init_depth_ablation}
\end{table}
\begin{table}[t]
    \centering
    \footnotesize
    \caption{
    Hybrid models ablation: Quantitative results with \textit{Protocol~1} on the \textit{Original} and \textit{Rotated} test set of Human3.6M.
    Hybrid 2 are models using the equivariant model first-layer features (see Figure~\ref{fig:hybrid_model_types}), resulting in the architecture change of the non-equivariant baseline and the corresponding increase of the parameter count (6.5M vs. 4.4M for VN-Transformer  (VNT) Hybrid, and 5.3M vs. 4.3M for GotenNet Hybrid).
    }
    \vspace{-5pt}
    \begin{tabularx}{\linewidth}{@{\extracolsep{\fill}}llcccc}
        \toprule
        \multirow{2}{*}{Method} & \multirow{2}{*}{Type} & \multicolumn{2}{c}{Human3.6M MPJPE ($\downarrow$)} 
        \\
        & & \textit{Original} & \textit{Rotated} 
        \\
        \midrule
        VNT & Hybrid & ${64.2}\pm{{0.1}}$ & $176.8\pm{1.0}$ 
    \\
        VNT & Hybrid 2& ${64.2}\pm{{0.4}}$ & $170.2\pm{1.1}$ \\
        VNT+aug & Hybrid & $66.0\pm{0.4}$ & $66.0\pm{0.3}$ \\ 
        VNT+aug & Hybrid 2& $66.1\pm0.7$ & $66.1\pm0.6$ \\
        \midrule
        GotenNet & Hybrid & $64.3\pm{0.3}$ & $177.5\pm{0.8}$ \\
        GotenNet & Hybrid 2 & $64.2\pm0.2$ & $166.2\pm0.5$ \\
        GotenNet+aug & Hybrid & $65.8\pm{0.2}$ & $65.9\pm{0.3}$ \\
        GotenNet+aug & Hybrid 2 & $66.0\pm0.3$ & $66.0\pm0.3$ \\
        \bottomrule
    \end{tabularx}
    \vspace{-25pt}
    \label{tab:hybrid_ablation}
\end{table}
\begin{figure*}[htbp]
    \centering
    \includegraphics[width=0.9\linewidth]{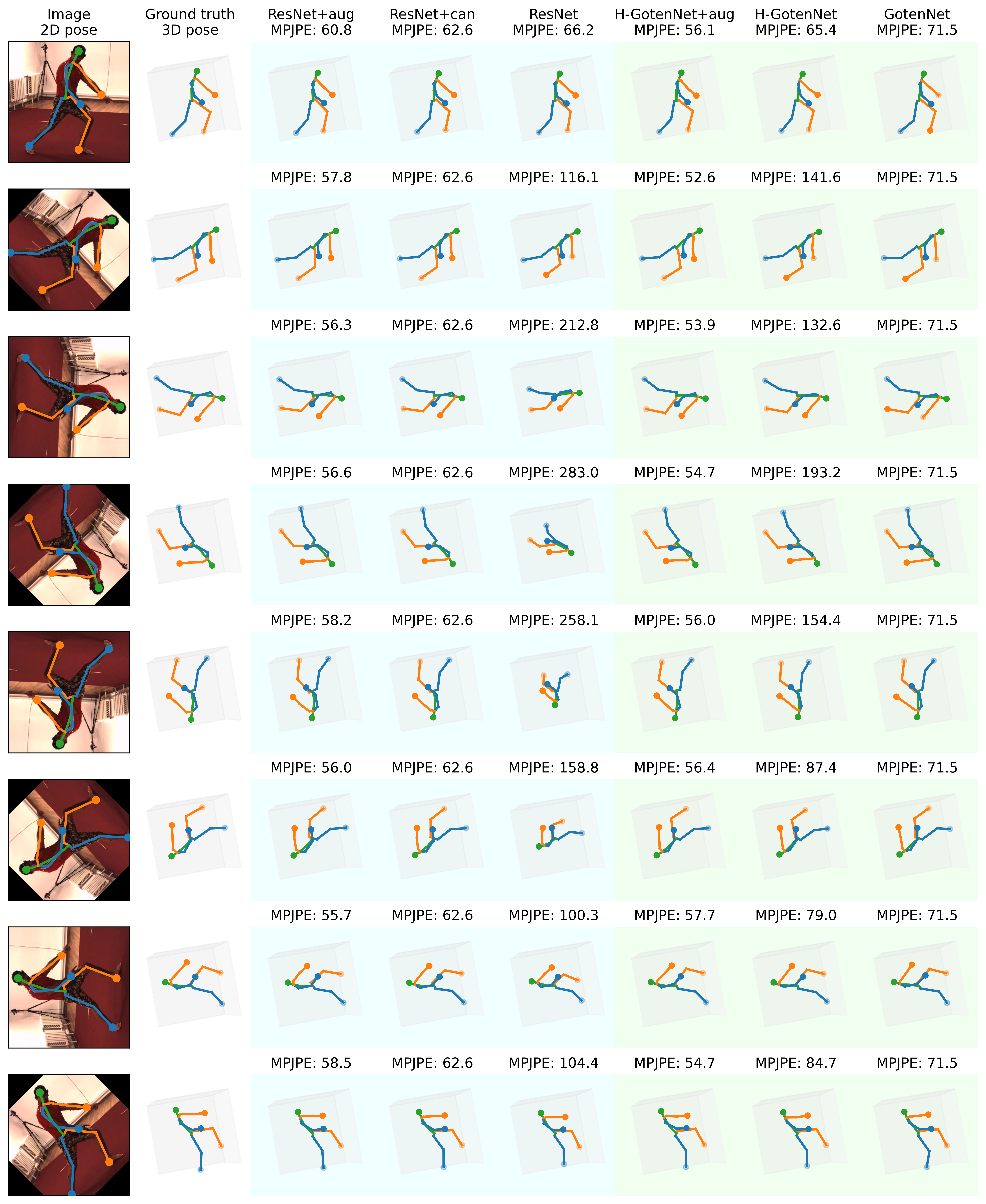}
    \vspace{-4pt}
    \caption{Qualitative rotation robustness comparison on Human3.6M. 
    Each row shows a test sample with different in-plane rotations of the input 2D pose (leftmost column).
    The corresponding 3D predictions are shown for different models.
    H-GotenNet denotes hybrid GotenNet.
    }
    \label{fig:rotations}
    \vspace{-5pt}
\end{figure*}
\section{Ablation studies}
\label{supp:ablations}
\vspace{-2pt}
\paragraph{Input depth initialisation for fully equivariant models}
In Table~\ref{tab:equi_init_depth_ablation}, we present an ablation study on the input depth initialisation for the fully equivariant models.
Here, we compare the selected initialisation with the norm of the 2D coordinates of each joint with other approaches. 
Namely, one alternative is when we append to every $N\times 2$ input a vector of size $N$, the elements of which are sampled independently from $\mathcal{N}(0, 0.1^2)$ when the model is instantiated (therefore, having the same specific depth per joint for all of the inputs), which we call \emph{fixed rand init $z$}.
Its modification is when this vector becomes learnable (\emph{learn init} $z$).
As we see from the results, making the initial input depth learnable provides no consistent improvement over fixed, randomly initialised depth. 
Similarly, the homogeneous input strategy (applicable only to the VN-Transformer) does not improve the results.
Hence, we use the norm depth initialisation ($z=\sqrt{x^2 + y^2}$) for both models.
%

\noindent\textbf{Hybrid model ablation}\quad We also conduct an ablation study for the hybrid models by using the first layer equivariant model features for the depth prediction (see Figure~\ref{fig:hybrid_model_types}).
The results presented in Table~\ref{tab:hybrid_ablation} suggest that no consistent improvement is gained compared to the original hybrid models, despite the parameter count increase caused by the growth of the dimensionality in the input to the non-equivariant baseline for depth prediction.
\section{Additional qualitative results}
\label{supp:additional_qual}
\begin{figure}
    \centering
    \includegraphics[width=\linewidth]{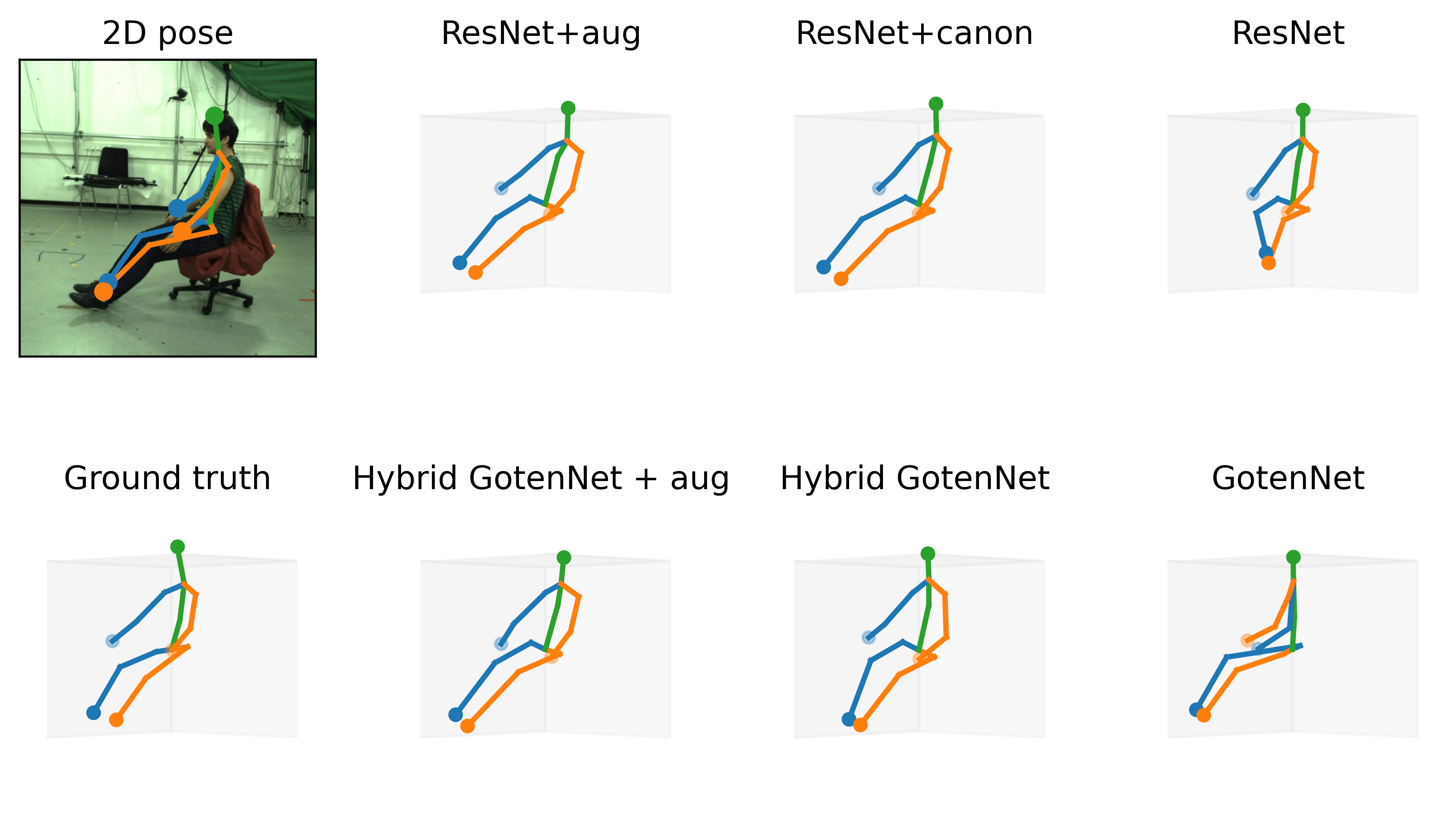}
    \caption{
    An example of the performance of different models on an MPI-INF-3DHP sample.
    All 3D estimation results are captured from the same viewing angle.
    }
    \label{fig:qual_3dhp}
\end{figure}

\begin{figure}[t]
    \centering
    \includegraphics[width=\linewidth]{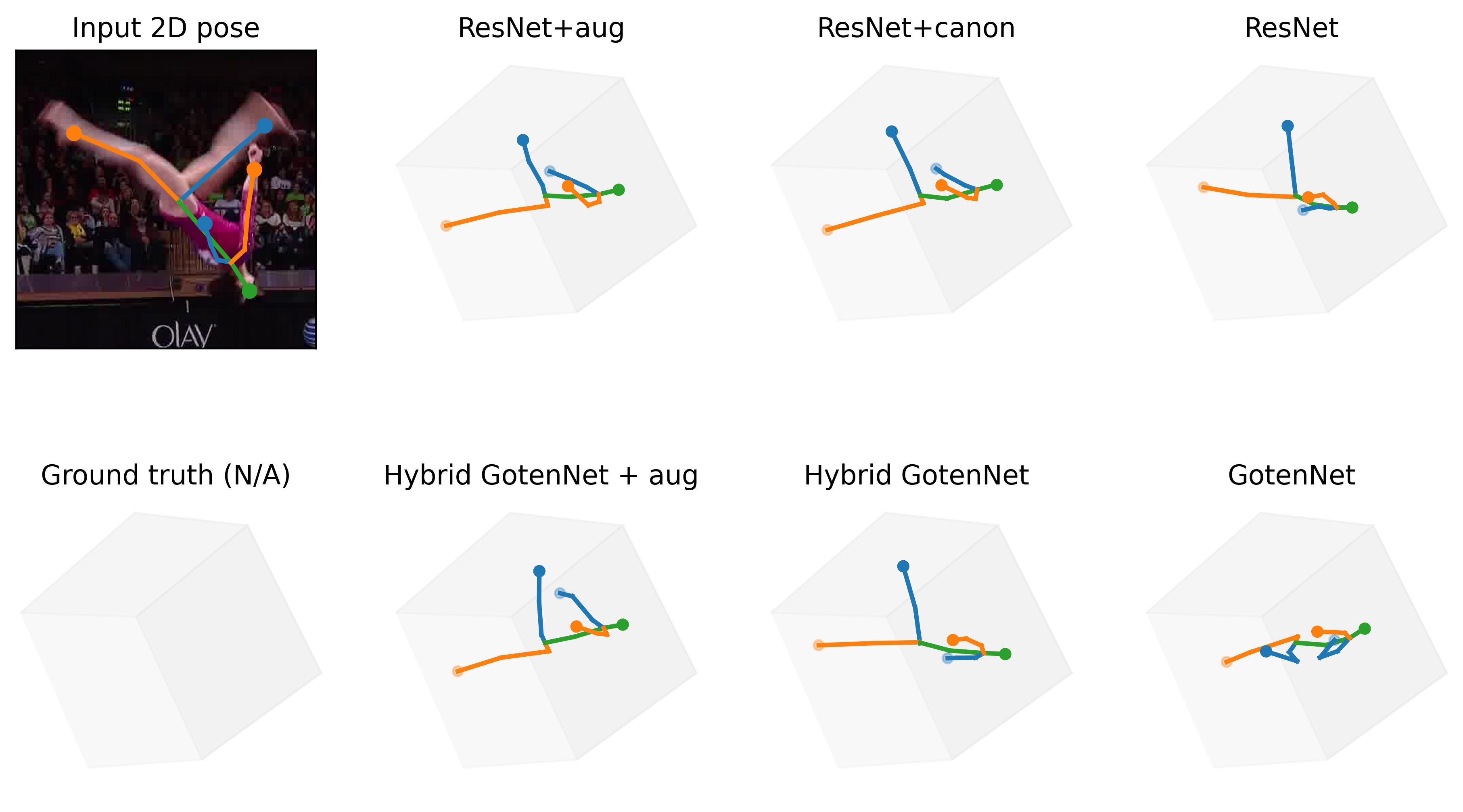}
    \\
    \vspace{10pt}
    \includegraphics[width=\linewidth]{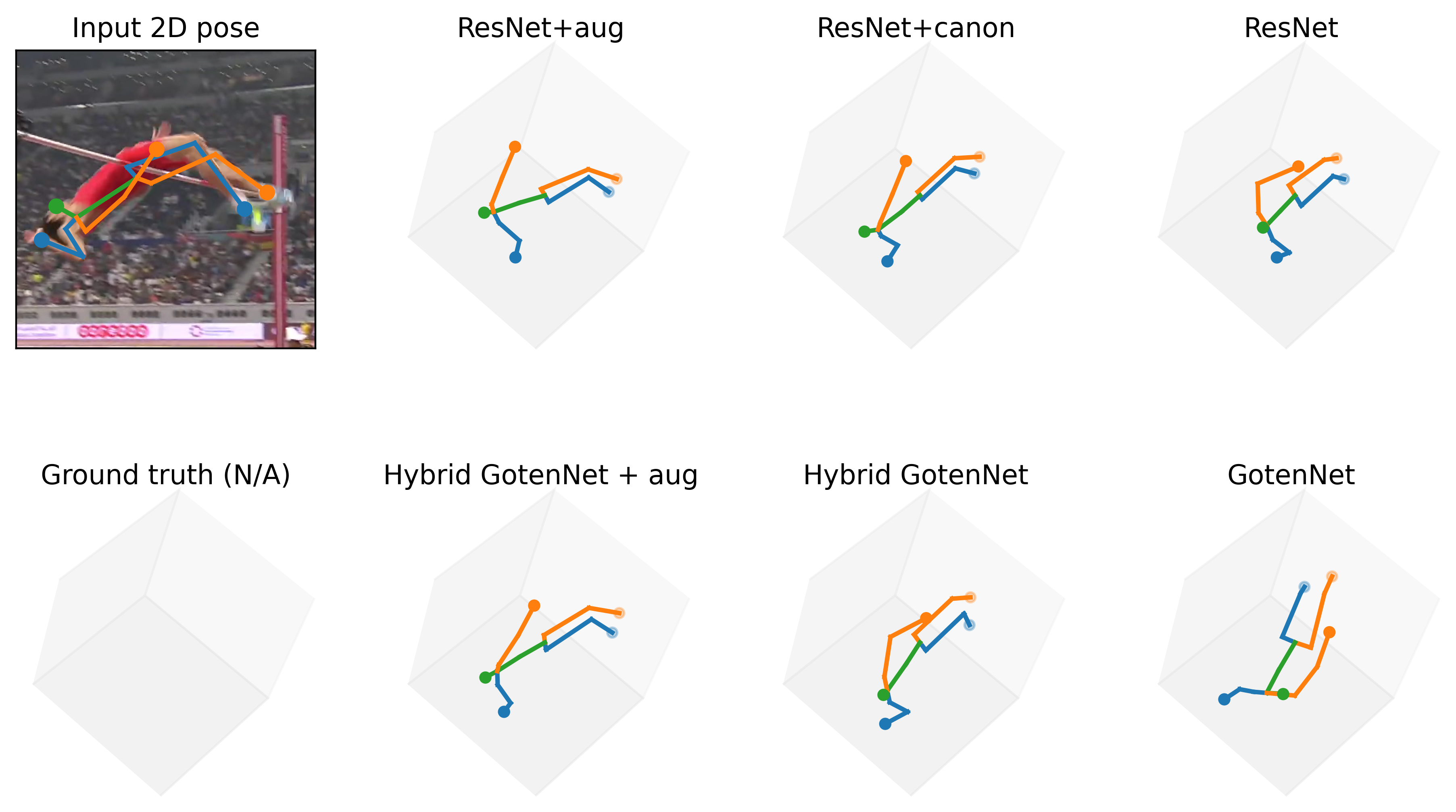}
    \caption{
    Representative examples of the performance of different lifting models on SportsCap samples (3D ground truth is unavailable) . 
    Ground-truth 2D keypoints are used as input to the lifting models, as the trained HRNet performs poorly on these SportsCap samples.
    All 3D estimation results are captured from the same viewing angle.
    }
    \label{fig:qual_sportscap}
\end{figure}

\begin{figure}[b]
    \centering
    \includegraphics[width=\linewidth]{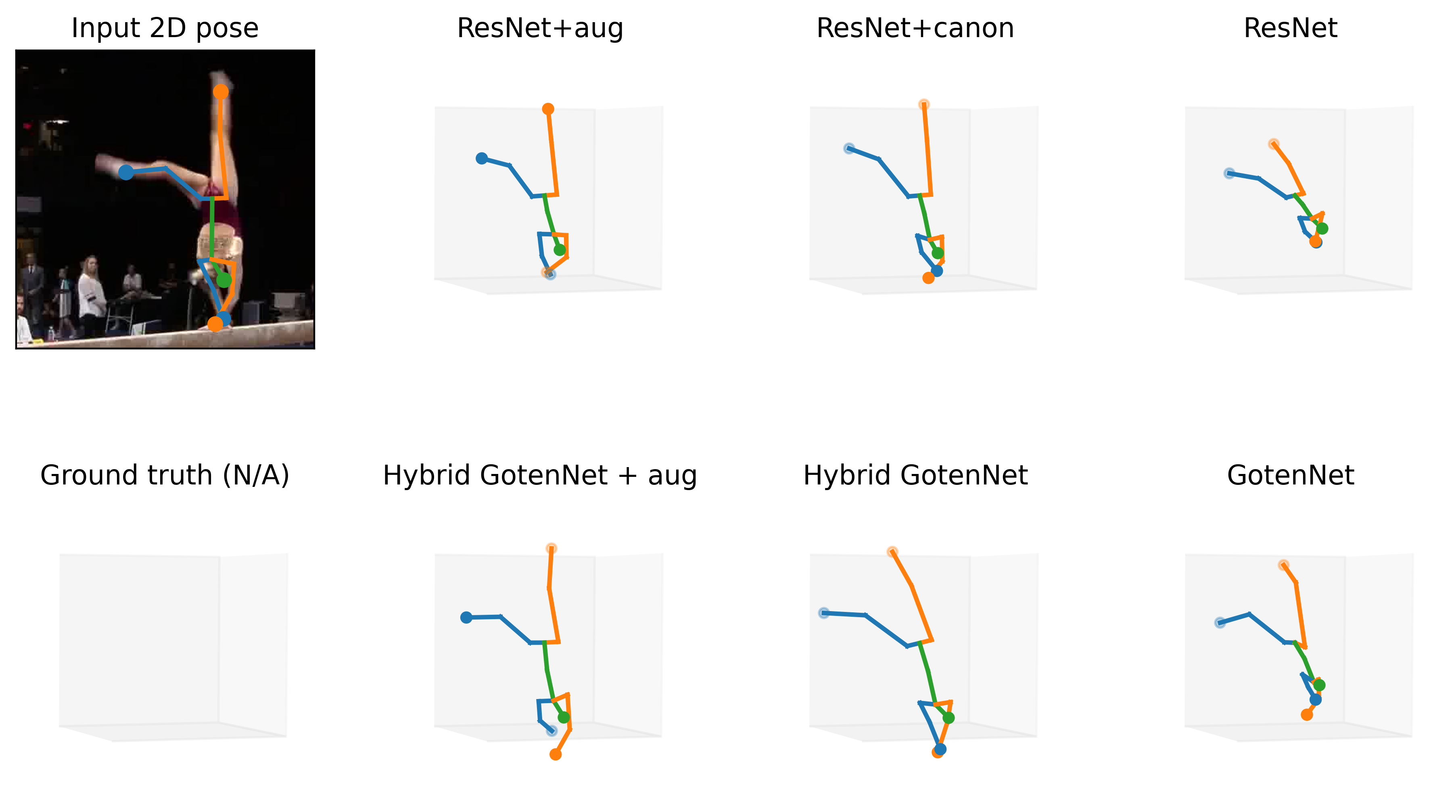}
    \\
    \vspace{10pt}
    \includegraphics[width=\linewidth]{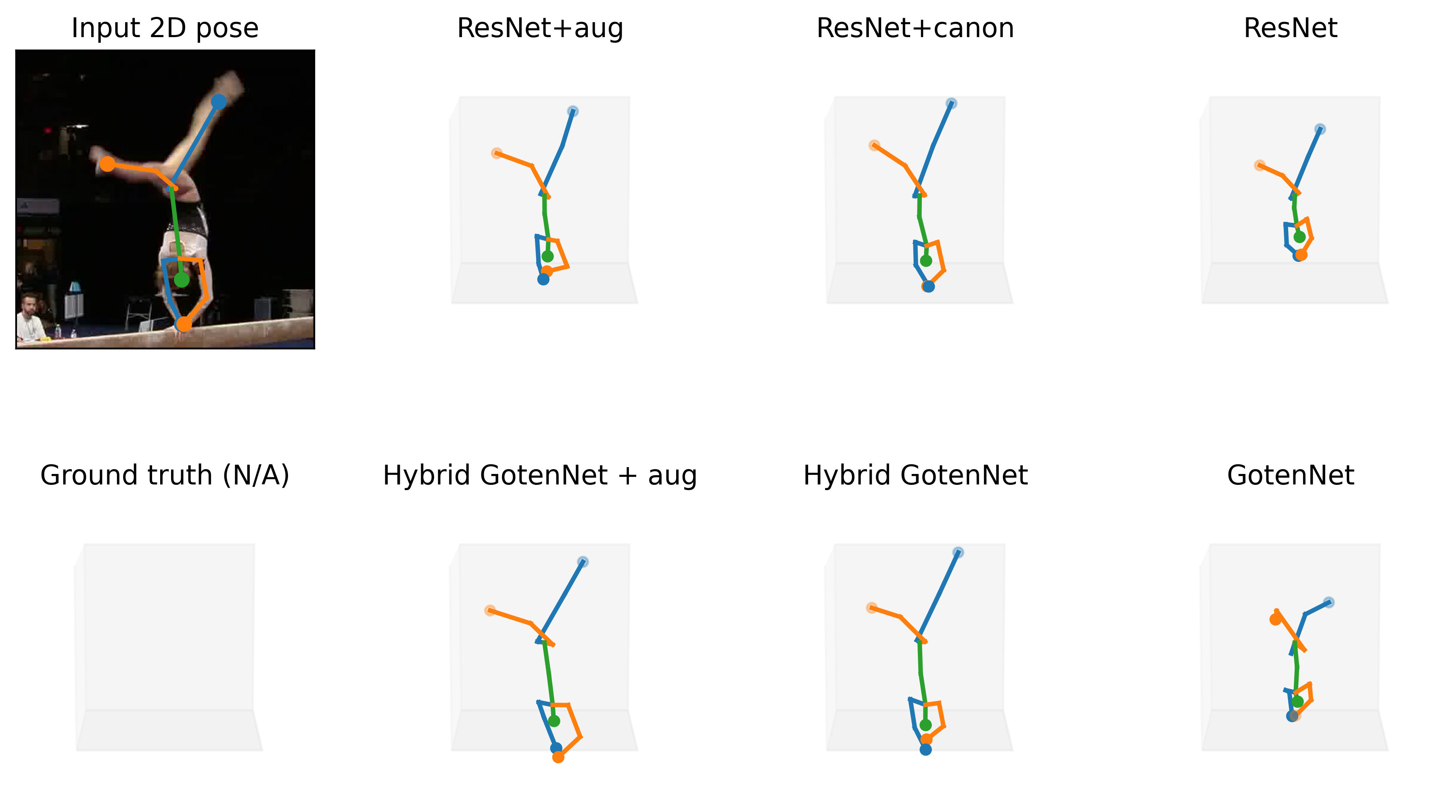}
    \\
    \vspace{10pt}
    \includegraphics[width=\linewidth]{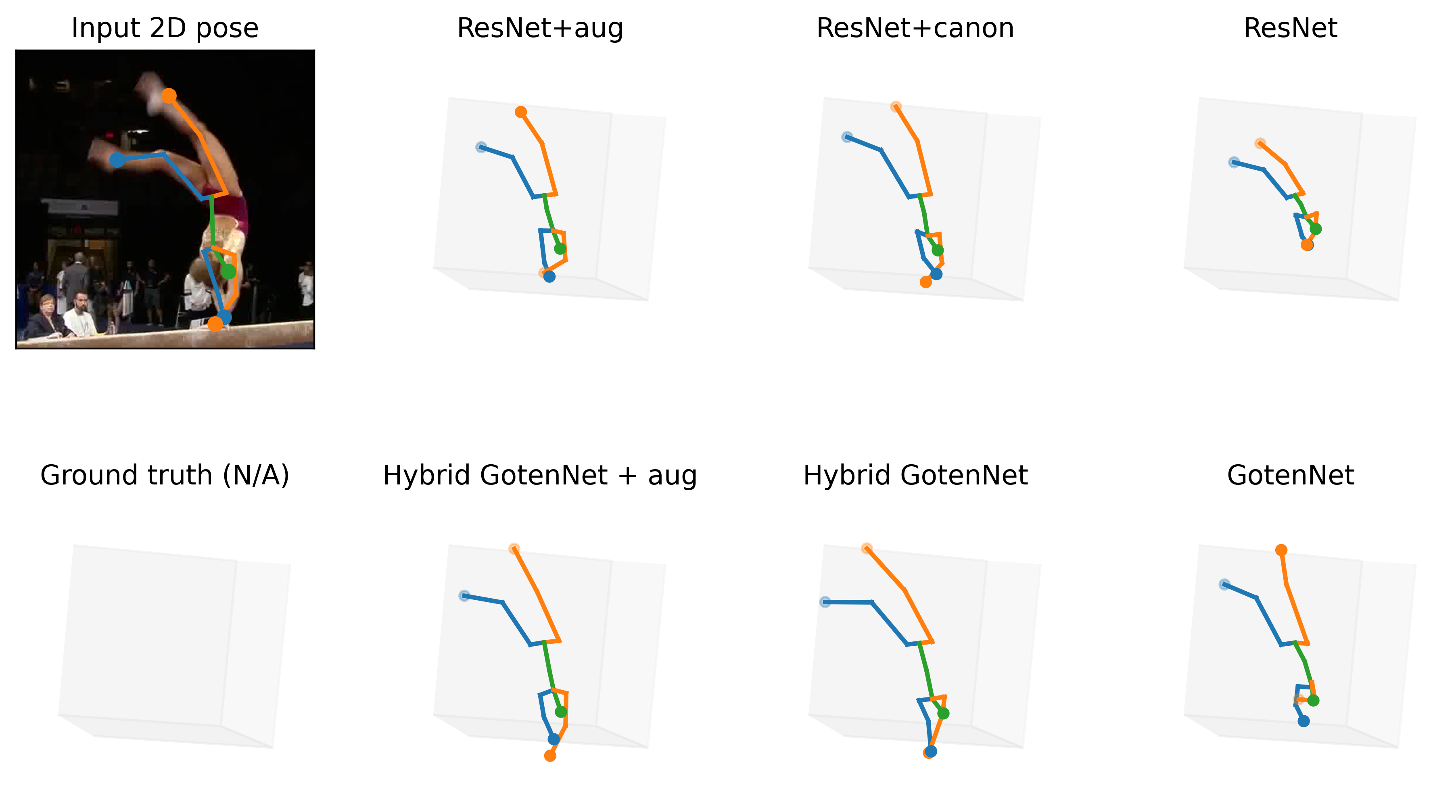}
    \caption{
    Examples of gymnastic poses from SportsCap.
    }
    \label{fig:qual_sportscap_gym}
\end{figure}

We additionally illustrate the rotation diagnostics with a qualitative example in Figure~\ref{fig:rotations}.
The performance of the lifting models on an MPI-INF-3DHP sample is presented in Figure~\ref{fig:qual_3dhp}.
Qualitative results on the gymnastic poses from the SportsCap dataset are demonstrated in Figures~\ref{fig:qual_sportscap}~and~\ref{fig:qual_sportscap_gym}.

\end{document}